\documentclass{article}
\PassOptionsToPackage{hidelinks}{hyperref}

 \usepackage[preprint]{neurips_2026}

\usepackage[utf8]{inputenc} 
\usepackage[T1]{fontenc}    
\usepackage{hyperref}       
\usepackage{url}            
\usepackage{booktabs}       
\usepackage{amsfonts}       
\usepackage{nicefrac}       
\usepackage{microtype}      
\usepackage{xcolor}         
\usepackage{graphicx}
\usepackage{enumitem}
\usepackage{amsmath}
\usepackage{array}
\usepackage{wrapfig}
\usepackage{floatflt}
\usepackage{lipsum}
\usepackage{booktabs}
\usepackage{multirow}
\usepackage{makecell}
\usepackage{xspace}
\usepackage{pifont}
\usepackage{colortbl}
\usepackage{booktabs}
\usepackage{makecell}
\usepackage{array}
\usepackage{pifont}
\usepackage{xcolor}
\usepackage{booktabs}
\usepackage{pifont}   
\usepackage{booktabs}
\usepackage{fontawesome5}
\usepackage{amssymb}
\usepackage{graphicx}
\usepackage{makecell}
\usepackage{booktabs}
\usepackage[table]{xcolor}
\usepackage{colortbl}
\usepackage{natbib}
\usepackage{tabularx}
\usepackage{graphicx}
\usepackage{booktabs}
\usepackage{subcaption}

\usepackage[most]{tcolorbox}
\usepackage{xcolor}

\usepackage{wrapfig}
\usepackage{fancyvrb}

\usepackage{longtable}

\definecolor{promptblue}{RGB}{235,244,255}
\definecolor{promptborder}{RGB}{60,110,180}
\definecolor{promptgray}{RGB}{248,248,248}
\newcommand{\cmk}{\textcolor{green!60!black}{\ding{51}}}
\newcommand{\xmk}{\textcolor{red!80!black}{\ding{55}}}

\usepackage{xcolor}
\usepackage{booktabs}
\usepackage{tabularx}
\usepackage[normalem]{ulem}
\usepackage{fancyvrb,fvextra}

\usepackage[most]{tcolorbox}
\usepackage{graphicx}
\usepackage{booktabs}

\usepackage[table]{xcolor}
\usepackage{soul}
\usepackage{tabularx}
\usepackage{booktabs}
\usepackage{makecell}

\PassOptionsToPackage{hidelinks}{hyperref}
\makeatletter
\newcommand{\symbolfootnotetext}[2][\textsection]{%
  \begingroup
    \c@footnote\z@
    \renewcommand\thefootnote{#1}%
    \stepcounter{footnote}%
    \protected@xdef\@thefnmark{\thefootnote}%
    \@footnotetext{#2}%
  \endgroup
  \setcounter{footnote}{0}%
}
\makeatother

\newcommand{\unitone}[1]{\sethlcolor{green!25}\hl{#1}}
\newcommand{\unittwo}[1]{\sethlcolor{cyan!25}\hl{#1}}
\newcommand{\unitthree}[1]{\sethlcolor{yellow!35}\hl{#1}}
\newcommand{\unitfour}[1]{\sethlcolor{orange!25}\hl{#1}}
\newcommand{\unitfive}[1]{\sethlcolor{purple!20}\hl{#1}}
\newcommand{\unitsix}[1]{\sethlcolor{red!20}\hl{#1}}

\definecolor{famP}{HTML}{DCEEFF}
\definecolor{famD}{HTML}{E8E0FF}
\definecolor{famM}{HTML}{E0F6EA}
\definecolor{famR}{HTML}{FFF0D6}
\definecolor{famW}{HTML}{FFE1E8}

\newtcolorbox{promptbox}[2][]{
    enhanced,
    colback=promptblue,
    colframe=promptborder,
    coltitle=black,
    fonttitle=\bfseries,
    title=#2,
    boxrule=0.6pt,
    arc=2mm,
    left=1.2mm,
    right=1.2mm,
    top=1mm,
    bottom=1mm,
    before skip=0.8em,
    after skip=0.8em,
    #1
}

\hypersetup{
    colorlinks=true,
    linkcolor=blue,
    citecolor=blue,
    filecolor=blue,
    urlcolor=blue,
}

\title{\dataset: Scientific Reasoning Supervision for Medical Vision–Language Models }

\author{%
  \textbf{Negin Baghbanzadeh}\textsuperscript{\textnormal{1,2\,\S}},
  \textbf{Pritam Sarkar}\textsuperscript{\textnormal{2,3}},
  \textbf{Michael Colacci}\textsuperscript{\textnormal{4,5}},
  \textbf{Abeer Badawi}\textsuperscript{\textnormal{1,2}}, \\[3pt]
  \textbf{Adibvafa Fallahpour}\textsuperscript{\textnormal{2,4,6,7}},
  \textbf{Arash Afkanpour}\textsuperscript{\textnormal{2}},
  \textbf{Leonid Sigal}\textsuperscript{\textnormal{2,3}},
  \textbf{Ali Etemad}\textsuperscript{\textnormal{8}},\\[3pt]
  \textbf{Elham Dolatabadi}\textsuperscript{\textnormal{1,2}} \\[15pt]
  \normalfont
  \textsuperscript{1}York University
  \textsuperscript{2}Vector Institute
  \textsuperscript{3}University of British Columbia\\
  \textsuperscript{4}University of Toronto
  \textsuperscript{5}Unity Health Toronto / St.\ Michael's Hospital\\
  \textsuperscript{6}University Health Network
  \textsuperscript{7}Arc Institute
  \textsuperscript{8}Queen's University
}

\newcommand{\negin}[1]{{\color{orange}[NB: #1]}}

\newcommand{\ps}[1]{{\color{blue}[PS: #1]}}

\newcommand{\dataset}{\textsc{OpenMedReason}\xspace}

\newcommand{\gainbase}{{\textsc{$20\%$}\xspace}}

\newcommand{\gainFM}{{\textsc{$4.2\%$}\xspace}}

\newcommand{\specialcell}[2][c]{%
\begin{tabular}[#1]{@{}c@{}}#2\end{tabular}}

\begin{document}

\maketitle
\symbolfootnotetext[\textsection]{Correspondence to: \href{mailto:negin.baghbanzadeh@vectorinstitute.ai}{neginb@yorku.ca}}

\begin{abstract}
High-stakes clinical use of large vision–language models (LVLMs) requires reasoning that is grounded in visual evidence and clinical knowledge, not just correct final answers. We introduce \dataset, a large-scale, open multimodal medical reasoning corpus comprising approximately 450K image–question–answer instances whose reasoning traces are primarily derived from curated biomedical, human-authored scientific articles. \dataset provides high-fidelity supervision beyond synthetic chains of thought, covering diverse medical domain vision modalities such as radiological scans, microscopic images, visible light photographs, charts, and others. We complement it with \dataset-Bench, a held-out benchmark that allows fine-grained evaluation of LVLMs along three complementary axes of capability, including perception, medical knowledge, and rationale, enabling diagnostic evaluation beyond final-answer accuracy. \dataset~is a rich training resource that exhibits its effectiveness in both supervised fine-tuning (SFT) and reinforcement-based alignment. Training with \dataset~yields a \gainbase~average improvement in VQA accuracy over the base model and achieves performance within \gainFM~of the strongest comparable-scale medical LVLMs. Fine-grained performance analysis confirms that the gains are not concentrated in any single axis: \dataset~improves perception, medical knowledge, and rationale jointly, and its reasoning traces are preferred over those of the base model in 86.1\% of pairwise comparisons. We release the code and dataset at 
\href{https://huggingface.co/datasets/neginb/OpenMedReason}
{\nolinkurl{huggingface.co/datasets/neginb/OpenMedReason}} .

\end{abstract}

\section{Introduction}

\begin{figure}[htbp]
    \centering
    \includegraphics[width=\textwidth]{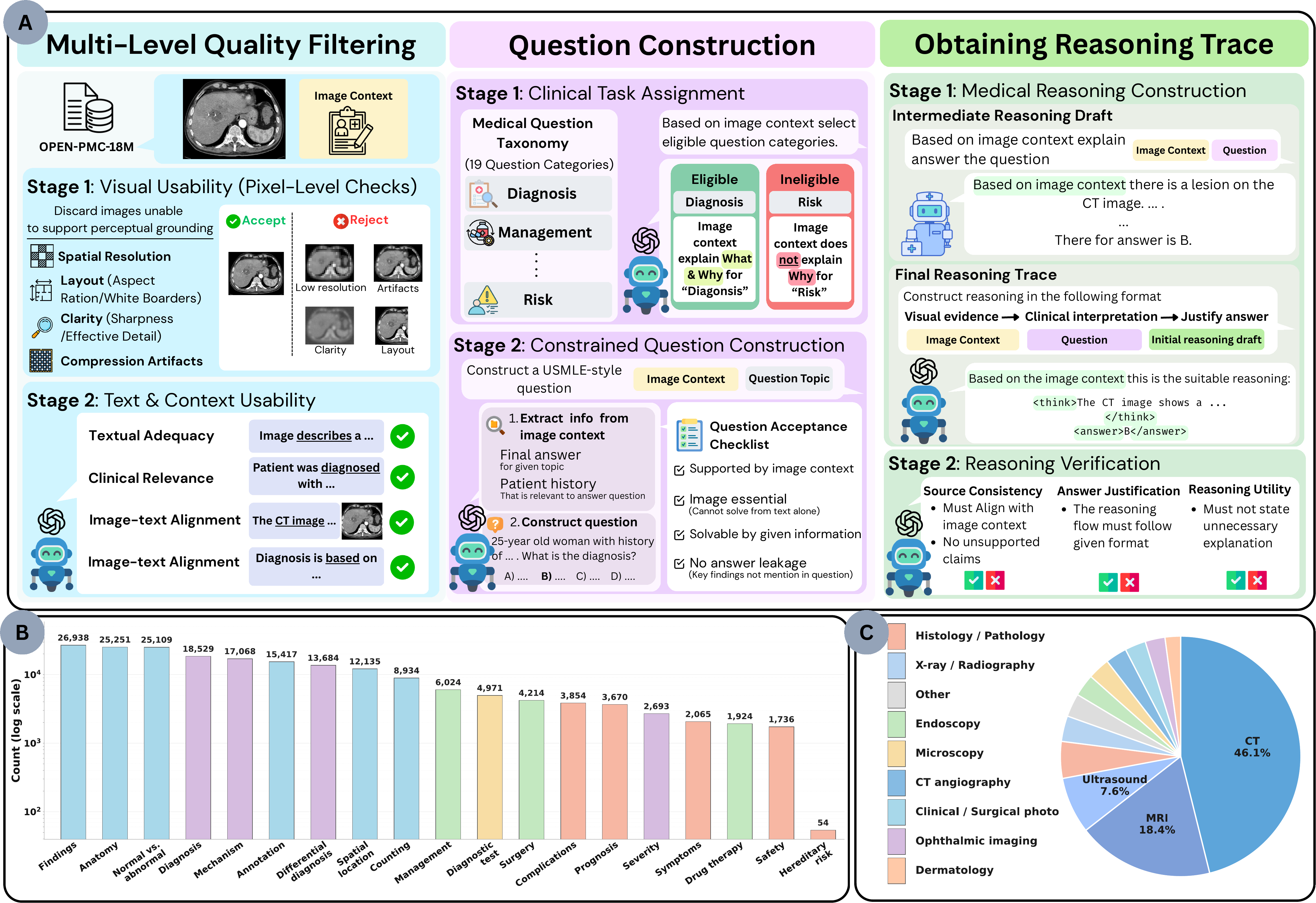}
    \caption{(A) Overview of the multi-stage \dataset curation pipeline, including quality filtering, context extraction, question construction, reasoning-trace generation, and verification. (B) Distribution of the 19 clinical task categories. Color coding corresponds to the task families defined in Table~7. (C) Distribution of imaging modalities in \dataset.}

    \label{fig:pipeline}
\end{figure}


Recent advances in large vision–language models (LVLMs) are reshaping how AI is applied in healthcare~\cite{li2023llava,sellergren2025medgemma,xu2025lingshu}. These models can interpret medical images, assist with report generation, and support diagnostic and prognostic decisions~\citep{sellergren2025medgemma,medpalm,baghbanzadehadvancing}. Clinical use, however, demands transparency: the model must make explicit how it arrived at its answer~\cite{gu2025illusion}. Outputs without a coherent rationale are difficult to trust, audit, or incorporate into clinical workflows \cite{jin2024hidden}. This places \emph{medical reasoning}, extracting the right visual evidence, invoking the relevant clinical knowledge, and connecting them through a valid inferential chain, at the center of clinical utility \cite{daiqoq}. Progress should therefore be measured not only by whether a model reaches the correct conclusion, but by whether the reasoning that produced it is one a clinician could follow and defend.

Recent reasoning models show that RL-centered post-training, often combined with supervised cold-start or distillation stages, can improve reasoning behavior \cite{guo2025deepseek}. However, these gains require reasoning supervision that is grounded in visual and clinical evidence and that captures the intermediate decisions a clinician would make~\cite{jeddi2026doesrlhelpmedical}. Despite its importance, open, medical reasoning supervision remains limited. Many existing resources are closed or designed primarily for benchmarking, while most public corpora provide only image–question–answer triples without reasoning traces. The few resources that do include traces often rely heavily on end-to-end LLM rollouts \cite{daiqoq,ossowski2025octomed}. While scalable, such traces inherit the priors and stylistic regularities of the generating model, narrowing the diversity and fidelity of the resulting supervision. As a consequence, supervised fine-tuning (SFT) may learn brittle reasoning patterns, while RL has little principled structure to refine. Motivated by this limitation, we ask the following question: \emph{To what extent does the effectiveness of an SFT$\rightarrow$RL pipeline in improving reasoning depend on the availability of high-quality reasoning-grounded supervision?}

We address this gap with \dataset, a large-scale, open multimodal medical reasoning resource explicitly designed to \emph{augment} existing datasets. \dataset comprises approximately \texttt{450K} image--question--answer instances, anchored by a core subset of \texttt{196K} examples derived from the Open-PMC corpus and paired with source-grounded reasoning traces. Spanning 19 clinical taxonomies and more than 8 imaging modalities, \dataset is, to our knowledge, among the largest openly released corpora whose reasoning traces are not reducible to a single model’s priors, but instead are anchored in scientifically documented visual evidence and clinical knowledge. Its taxonomy spans interpretation, diagnosis, clinical decision-making, risk and prognosis, and mechanistic reasoning, providing broad supervision for SFT and a structured foundation for subsequent RL-based alignment. We train an SFT$\rightarrow$GRPO pipeline using \dataset and benchmark it against strong open-source and domain-specific medical models across 14 benchmarks. As a step toward clinically auditable evaluation, we pair these results with a diagnostic framework that, for the first time, decomposes model behavior into \textit{\textbf{perception}}, \textit{\textbf{knowledge}}, and \textit{\textbf{rationale}} contributions, and we release the trained model so the pipeline can be stressed beyond our own evaluations.


Our contributions are as follows:

{\large\ding{192}}~
\textbf{\dataset (a reasoning-supervision source).} A large-scale, open multimodal medical reasoning corpus whose core reasoning traces are derived from curated, human-authored scientific context and not end-to-end LLM rollouts. \dataset's richness and clinical coverage allow it to be combined with existing datasets to support both SFT-based reasoning training and RL-based alignment.

{\large\ding{193}}~    
\textbf{\dataset-Bench (a capability-level benchmark).} A held-out evaluation suite that decomposes multimodal medical reasoning into three distinct capabilities, perception, medical knowledge, and reasoning, which enables capability-resolved diagnosis of model behavior beyond final-answer accuracy.

{\large\ding{194}}~
\textbf{Empirical evidence for grounded supervision in medical LVLM post-training.} Through experiments on 14 medical benchmarks, we demonstrate that source-grounded reasoning supervision improves SFT→RL post-training in medical LVLMs. \dataset improves the model’s ability to surface clinically relevant visual and contextual evidence and increases the diversity of correct reasoning paths explored during decoding.



\section{Related Work}

\textbf{Multimodal Foundation Models for Medicine.}
\label{sec:rw-models}
Medical LVLMs have evolved from general-purpose multimodal backbones adapted to biomedical data toward domain-specialized systems trained on increasingly large and heterogeneous medical corpora~\citep{li2023llava}. More recent open efforts, including OctoMed~\citep{ossowski2025octomed}, MedGemma~\cite{sellergren2025medgemma}, Lingshu~\cite{xu2025lingshu}, MedVL-Thinker~\citep{huangmedvlthinker}, and QoQ-Med~\citep{daiqoq}, scale this paradigm with different backbones, modality coverage, and data construction strategies (Table~\ref{tab:related_work}(a)).



\textbf{Medical VQA datasets and benchmarks.}
\label{sec:rw-data}
Early datasets such as, VQA-RAD~\citep{lau2018dataset}, SLAKE~\citep{liu2021slake}, PathVQA~\citep{he-etal-2021-towards}, PMC-VQA~\citep{zhang2024development}, and OmniMedVQA~\cite{hu2024omnimedvqa} together with the more challenging expert-level MedXpertQA-MM~\citep{zuo2025medxpertqa} and JAMA CC-MM~\cite{ama_jama_challenge_2024}, primarily score models by final-answer correctness. Large-scale biomedical image--text corpora such as Open-PMC~\citep{baghbanzadehadvancing} and Open-PMC-18M~\citep{baghbanzadeh2025open} provide broad coverage for representation learning but offer no reasoning supervision or capability-level evaluation. M3CoTBench~\citep{jiang2026m3cotbench} moves toward chain-of-thought evaluation in the medical domain, but does not separately attribute failures to perception, knowledge, or reasoning. \dataset-Bench is designed to fill this gap: each item is annotated for the cognitive requirements it exercises, and model traces are scored using a capability-specific rubric (Table~\ref{tab:related_work}(b)).

\textbf{Reasoning supervision and post-training.}
\label{sec:rw-reasoning-data}
Recent reasoning models show that supervised reasoning data and RL-based post-training can improve model behavior, particularly when RL is initialized from a strong supervised checkpoint~\citep{guo2025deepseek}. In the medical multimodal setting, methods such as Med-R1~\citep{lai2026med}, and QoQ-Med~\citep{daiqoq} explore reasoning-oriented post-training for medical VQA. Beyond medicine, BioReason~\citep{bioreason} and BioReason-Pro~\citep{bioreason-pro} extend multimodal reasoning to biology. These works suggest that RL can sharpen answer selection and improve reasoning-style outputs. Our work asks a prior question: what supervision should define the model's reasoning behavior before RL is applied? \dataset addresses this gap by providing a plug-in reasoning resource for SFT and RL alignment, while \dataset-Bench evaluates whether models improve not only in answer accuracy but also in perception, medical knowledge, and reasoning quality.

\begin{table*}[t]
\centering
\caption{Comparison of multimodal medical resources. \textbf{(a) Training datasets}- Reas.: includes reasoning traces. New Qs: introduces new questions. Grounded Traces: reasoning traces grounded in scientific evidence rather than solely LLM-generated. \textbf{(b) Evaluation benchmarks}- Metrics: A=final Accuracy, F1=F1-score, BLEU=BLEU score, Reas. Eval: evaluates intermediate capabilities i.e., perception (P), knowledge (K), and reasoning (R), beyond final answers. "Mixed" refers to multiple modalities. 
}
\label{tab:related_work}
\setlength{\tabcolsep}{3pt}

\begin{minipage}[t]{0.48\textwidth}
\centering
{\small \textbf{(a) Training datasets}}\\[2pt]
\renewcommand{\arraystretch}{1.4}  
\resizebox{\linewidth}{!}{%
\begin{tabular}{l c c c c c}
\toprule
\textbf{Dataset} & \textbf{\#Samples} &
\textbf{Reas.} &
\makecell{\textbf{New}\\\textbf{Qs}} &
\makecell{\textbf{Grounded}\\\textbf{Traces}} &
\makecell{\textbf{Open}\\\textbf{Src.}} \\
\midrule
LLaVA-Med~\cite{lau2018dataset}          & 60k   & \xmk & \cmk & \xmk & \cmk \\
Med-PaLM M~\cite{medpalm}                & 1M    & \xmk & \xmk & \xmk & \xmk \\
MedGemma~\cite{sellergren2025medgemma}   & 1.6M  & \xmk & \xmk & \xmk & \xmk \\
PubMedVision~\cite{chen2024towards} & 1.3M  & \xmk & \cmk & \xmk & \cmk \\
CLIMB~\cite{dai2025climb}                & 2.6M  & \xmk & \xmk & \xmk & \cmk \\
OctoMed~\cite{ossowski2025octomed}       & 8M    & \cmk & \xmk & \xmk & \xmk \\
MedTrinity-25M~\cite{xie2024medtrinity}  & 25M   & \xmk & \cmk & \xmk & \cmk \\
PMC-VQA~\cite{zhang2024development}      & 227k  & \xmk & \cmk & \xmk & \cmk \\
\midrule
\textbf{\dataset (Ours)} & \textbf{450k} & \cmk & \cmk & \cmk & \cmk \\
\bottomrule
\end{tabular}%
}
\end{minipage}%
\hfill
\begin{minipage}[t]{0.50\textwidth}
\centering
{\small \textbf{(b) Evaluation benchmarks}}\\[2pt]
\renewcommand{\arraystretch}{1.3}
\resizebox{\linewidth}{!}{%
\begin{tabular}{l c c c c}
\toprule
\textbf{Benchmark} & \textbf{\#Samples} & \textbf{Metrics} & \textbf{Modalities} &
\makecell{\textbf{Reas.}\\\textbf{Eval}} \\
\midrule
SLAKE~\cite{liu2021slake}                & 1.1k  & A, BLEU & CT, MRI, X-ray & \xmk \\
VQA-RAD~\cite{lau2018dataset}            & 0.5k  & A       & Radiology      & \xmk \\
PathVQA~\cite{he-etal-2021-towards}      & 6.8k  & A, BLEU & Pathology      & \xmk \\
PMC-VQA~\cite{zhang2024development}      & 2k    & A       & Mixed      & \xmk \\
OmniMedVQA~\cite{hu2024omnimedvqa}       & 128k  & A       & Mixed     & \xmk \\
MedXpertQA-MM~\cite{zuo2025medxpertqa}   & 2k    & A       & Rad., Path.    & \xmk \\
JAMA CC-MM~\cite{ama_jama_challenge_2024}                 & 1.5k  & A       & Mixed          & \xmk \\
GMAI-MMBench~\cite{chen2024gmai}         & 25.7k & A       & Mixed          & \xmk \\
ProbMed~\cite{elallaf2026medprobclip}            & 57.1k & A       & CT, MRI, X-ray & \xmk \\
CARES~\cite{xia2024cares}                & 41k   & A, F1   & Mixed          & \xmk \\
\midrule
\textbf{\dataset-Bench (Ours)} & \textbf{1.5k} & \textbf{P, K, R, A} & \textbf{Mixed} & \cmk \\
\bottomrule
\end{tabular}%
}
\end{minipage}

\end{table*}

\section{\dataset Curation}

In this section, we describe the multi-stage curation pipeline for \dataset, a large-scale
multimodal dataset for training and evaluating clinically grounded reasoning
(Figure~\ref{fig:pipeline}). The core of \dataset is constructed from OpenPMC-18M \cite{baghbanzadeh2025open} approximately 18 million image-text pairs from
biomedical publications, whose figure-caption (subfigures and subcaptions) are extracted from biomedical publications. 
Three principles guide the core construction:
(\textit{i}) \emph{Clinical grounding}, where each question requires linking visual findings to clinical or biomedical context, rather than recognizing isolated labels.
(\textit{ii}) \emph{Multimodal coverage}, where the dataset spans radiology, microscopy, and visible-light photography, with clinical-task categories that go beyond diagnosis (Figure~\ref{fig:pipeline}(C)). (\textit{iii}) \emph{Reliable supervision}, where an instance is retained only when the image is interpretable, the answer is recoverable from the provided evidence, and a source-grounded reasoning trace can be verified.

\subsection{Multi-Level Quality Filtering}
\label{sec:quality-filtering}
Each raw instance in Open-PMC-18M consists of a subfigure image from biomedical publications, its caption, and the in-text references to that figure. Most raw pairs are unsuitable for reasoning-oriented VQA, so we apply a two-stage filtering before question generation, including a pixel-level visual-usability stage followed by an LLM-based text-and-context stage. 


\textbf{Visual Usability.}
We first discard images that cannot support reliable perceptual grounding, applying pixel-level checks for spatial resolution, layout (aspect ratio and border content), clarity (sharpness and effective detail), and compression artifacts. Details of the filtering procedure are provided in Appendix~\ref{app:visualfilter}, with examples shown in Figure~\ref{fig:visual_quality_filter_examples}. Running these checks before any semantic processing prevents downstream stages from generating questions whose answers depend on evidence that is absent or too degraded to interpret. We retain approximately 30K high-quality samples at the end of this stage. 


\textbf{Text and Context.}
Each selected image is paired with its caption and in-text references and assessed by an LLM-based quality filter on four criteria. (\textit{i}) \emph{Textual adequacy}: the context conveys concrete information about the specific subfigure; (\textit{ii}) \emph{Clinical relevance}: the case is clinically or biomedically meaningful, excluding schematics, charts, workflow diagrams, instruments, purely molecular figures, and animal-model studies; (\textit{iii}) \emph{Image--text alignment}: the text describes the visual content rather than a loosely related figure; and (\textit{iv}) \emph{Reasoning readiness}: the context contains enough evidence to support a question, its answer, and a verifiable rationale. Additional details about the filtration criteria, decision rules, manual-validation results, and rejected-case and accepted-case examples are deferred to Tables~\ref{tab:text_quality_rejected_examples} and \ref{tab:text_quality_accepted_examples} in Appendix~\ref{app:textfiltering}.



\subsection{Question Construction}

Each retained image--context pair is converted into one or more image-dependent clinical reasoning questions. A caption or paragraph that mentions a disease, procedure, or finding does not necessarily support a valid VQA item since the answer may be stated without explanation, the visual evidence may be incidental, or the question may be solvable from the text alone. We therefore perform the question construction in two separate stages. We first identify which clinical tasks each pair can support, and then generate questions for those tasks under constraints designed to preserve visual grounding, prevent answer leakage, and ensure the rationale can be derived from the source text.


\textbf{Clinical Task Assignment.}
We define a clinical-task taxonomy (Figure \ref{fig:pipeline}(B)) that captures the intended objective of each question (Appendix~\ref{app:clinicaltax}). We developed the taxonomy via an inductive, bottom-up analysis of a randomly sampled subset of questions, where two independent annotators iteratively grouped questions by underlying clinical intent and progressively merged semantically overlapping groupings into higher-level themes, arriving at $19$ clinically motivated categories organized into five broader families (full taxonomy in Table~\ref{tab:task-taxonomy} and category-level statistics in Figure~\ref{fig:benchmark_stats} in Appendix~\ref{fig:benchmark_stats}). Each pair is assigned to one or more categories by prompting an LLM (\texttt{gpt-5-mini}) with the image, its modality, and the associated textual context, and asking it to return all categories supported by the pair as a structured XML list (prompt in Appendix~\ref{app:clinicaltax}). The prompt requires that a category be selected only when the source context both (i)~contains a plausible final answer for the task and (ii)~provides sufficient supporting evidence to ground a reasoning trace, rather than merely stating the answer itself. For example, a sentence such as “the patient was diagnosed with X” may support answer extraction, but is excluded unless the surrounding text also explains the visual or clinical basis for the conclusion. 

\newcommand{\famtag}[2]{{\color{#1}\rule[-0.1em]{0.65em}{0.65em}}\,\textbf{#2}}


\textbf{Constrained Question Generation.}
For each eligible (image, context, task label) tuple, we use \texttt{gpt-5-mini} (\texttt{reasoning\_effort=medium}) to generate a USMLE-style \cite{usmle_overview} item consisting of a question stem, answer options, and the correct answer. Generation is posed as a constrained validity problem: an item is retained only if it satisfies five conditions designed to preserve grounded multimodal reasoning. Specifically, the question, options, answer, and rationale must all be supported by the original image–context pair; the item must be answerable without external information; the image must provide essential evidence such that the question cannot be solved from text alone; the stem must avoid explicit leakage of the diagnosis or key visual findings; and long contextual vignettes may provide clinical background but cannot describe the image evidence the model is expected to infer. Each accepted item is further routed into either a short-context, image-forward format or a long-context clinical vignette depending on the assigned task category and the level of contextual reasoning supported by the source. Perception-oriented categories typically use shorter stems, whereas decision-oriented categories use longer clinical narratives. Answers are generated using task-specific formats including multiple choice, \textsc{Yes/No}, \textsc{True/False}, and \textsc{Normal/Abnormal}. Finally, annotation/marker and spatial-localization questions receive additional anti-leakage safeguards, since these categories are especially sensitive to implicit disclosure of visual evidence in the stem. See Figure~\ref{fig:pipeline} for an example and Appendix~\ref{app:questiongen} for full details of the question-generation process. Table~\ref{tab:question_examples} in Appendix~\ref{app:QuestionExamples} provides representative examples for each question type.

\subsection{Reasoning Trace Supervision}
\label{sec:reasoning-generation}


\textbf{Medical Reasoning Generation.}
For each accepted question--answer pair, we obtain a scientifically grounded medical reasoning trace. The trace is designed to be source-grounded, with the accompanying image context serving as the primary source of evidence for the visual findings, clinical interpretation, and answer justification. The reasoning is anchored in the evidence provided by the source image context, ensuring consistency with the visual and clinical information associated with the case. Instances without image context are excluded from this stage. 
Our reasoning pipeline consist of a two-stage procedure.
First, a medical LVLM~\cite{ossowski2025octomed} produces a draft conditioned on the image, source context, question, and correct answer. This serves as an evidence-organization step, linking the relevant visual findings, the question, and supporting source statements into a coherent explanation. 
The draft is then refined with \texttt{gpt-5-mini}, which aligns each portion against the image, question, answer, and caption / in-text references, removes unsupported or article-level content, and rewrites the rationale into a fixed evidence-to-answer format: 
modality and visual target $\to$ visual evidence
$\to$ clinical interpretation $\to$ justified answer.
Each component in the reasoning chain serves a distinct inferential role. \emph{Modality and visual target} specifies the imaging modality and anatomical or pathological focus; \emph{visual evidence} captures the clinically relevant findings grounded in the image; \emph{clinical interpretation} integrates those findings with patient context and medical knowledge; and \emph{answer justification} connects the resulting interpretation to the final response. Additional details and examples are provided in Appendix~\ref{app:reasoning}.

\textbf{Final Reasoning Verification.}
\label{sec:reasoning-verification}
After reasoning generation, we apply a final verification step to assess whether each trace should be retained. The verification stage evaluates whether each generated reasoning trace satisfies three requirements for grounded supervision. First, the trace must remain consistent with the source: every claim should be supported by the question, the image, or the associated context, with no extraneous information introduced. Second, the trace must justify the final answer by showing how it follows from the visual and contextual evidence, rather than restating isolated facts. Third, the explanation must remain useful and answer-directed without drifting into generic background knowledge or irrelevant article-level discussion. If any of these criteria are not satisfied, the instance is discarded. This final verification step is performed using \texttt{gpt-5-mini}, which provides an additional safeguard against traces that are fluent but weakly grounded, overly generic, or misaligned with the evidence. See Figure~\ref{fig:pipeline} for an example and Appendix~\ref{app:reasoning-verification} for full details and prompts for reasoning verification. 


\subsection{\dataset-Bench: Reasoning Trace Evaluation}
\label{sec:reasoning-benchmark}

\paragraph{Reasoning-trace evaluation.}
In addition to the training dataset, we introduce \dataset-Bench for fine-grained evaluation of LVLM capabilities beyond final answer accuracy, with a particular focus on assessing whether the reasoning trace contains the components necessary to justify the answer.
For each example, we
use \texttt{gpt-5-mini} to convert the reference reasoning trace, together with the
question and answer, into a compact checklist of atomic unit questions. Each unit question is designed under one of three axes: \emph{perception}, which covers
image-grounded observations; \emph{medical knowledge}, which covers general clinical facts;
and \emph{reasoning}, which covers case-specific inferential links from the evidence to the
answer.

Newly generated traces are scored against this checklist using two independent probes.
\emph{Presence} measures whether the trace engages with the unit's topic at all, while
\emph{correctness} measures whether the unit claim is stated accurately. For each axis $a$, we compute a normalized
presence score $\mathrm{Presence}_a$ over all units and a correctness rate
$\mathrm{Correctness}_a$ over only the units that are present. The final
trace score is
\[
\mathrm{TraceScore}
=
\frac{1}{2}
\sum_{a \in \{\mathrm{perc}, \mathrm{know}, \mathrm{rat}\}}
\mathrm{Presence}_a \cdot \mathrm{Correctness}_a .
\]
We report this joint score together with the six axis-level metrics, allowing us to
diagnose whether failures arise from missing visual evidence, incorrect clinical knowledge,
or broken inferential links. Details of this metric are provided in Appendix~\ref{app:reasoning_trace_eval}.

\subsection{\dataset Composition}
To broaden instruction tuning beyond the OpenPMC-derived subset, we mix \dataset with existing medical VQA resources (see Table~\ref{tab:sft_training_data} in Appendix~\ref{app:train_detail} for a full list of datasets). We used \texttt{gpt-5-mini} to generate reasoning traces. 
This combines source-grounded reasoning supervision with broader medical VQA breadth. In total, \dataset contains approximately \texttt{450K} image--question--answer instances: \texttt{196k} OpenPMC-derived examples with curated reasoning traces, and \texttt{254k} auxiliary VQA-derived examples. Coverage spans the imaging modalities and clinical-task families described above. \dataset--Bench is a separate held-out split of \texttt{1.5k} samples reserved exclusively for evaluation (Section~\ref{sec:reasoning-benchmark}).

\section{Experiments}

\begin{table}[t]
\centering
\caption{\textbf{Ablation of \dataset and post-training methods.} 
We report accuracy (\%) on five in-distribution (ID) and two out-of-distribution (OOD)
benchmarks, together with their group-wise means (ID~Avg, OOD~Avg) and the overall mean (Avg). Green and blue cell shading scale with the magnitude of improvement over the base model on ID and OOD averages, respectively. Bold marks the best result per group.}
\setlength{\tabcolsep}{3pt} 
\renewcommand{\arraystretch}{1.1}
\resizebox{0.99\linewidth}{!}{
\begin{tabular}{lcccccc|ccc|c}
\toprule
\textbf{Training Setup} 
& \specialcell{\bf \dataset\\\bf -Bench (Ours)}
& \specialcell{\bf SLAKE\\\cite{liu2021slake}} 
& \specialcell{\bf VQA\\\textbf{Rad} \cite{lau2018dataset}} 
& \specialcell{\textbf{Path}\\ \textbf{VQA} \cite{he-etal-2021-towards}} 
& \specialcell{\bf PMC\\\bf VQA  \cite{zhang2024development}} 
& \specialcell{\bf ID\\\bf Avg} 
& \specialcell{\bf MedXpert\\\textbf {QA} \cite{ding2026mmedexpert}}
& \specialcell{\bf JAMA\\\cite{ama_jama_challenge_2024}} 
& \specialcell{\bf OOD\\\bf Avg}
& \textbf{Avg} \\
\midrule
Qwen2.5-VL-7B (Base)
& 47.11 & 62.26 & 67.13 & 62.88 & 49.6 & \cellcolor{green!5}60.47 & 22.62 & 35.34&\cellcolor{blue!5}28.98 & \cellcolor{purple!5}53.10 \\
+ SFT on \dataset w/o Open-PMC 
& 48.20 & 79.57 & 69.32 & 62.37 & 54.31 &\cellcolor{green!15}66.39& 22.70 & 36.92&\cellcolor{blue!15}29.81 & \cellcolor{purple!15}57.65 \\
+ SFT on \dataset 
& 77.39 & 80.52 & 70.51 & 63.50 & 55.05 &\cellcolor{green!29}67.40& 26.16 & 38.42&\cellcolor{blue!28}32.29 & \cellcolor{purple!29}59.14 \\
\midrule
+ SFT + GRPO on \dataset
& 78.51 & 85.10 & 72.51 & 64.10 & 55.20 & \cellcolor{green!38} \textbf{69.23} & 24.95 & 39.97 & \cellcolor{blue!38} \textbf{32.46} & \cellcolor{purple!38} \textbf{60.04} \\
\bottomrule
\end{tabular}
}
\label{tab:increment}
\end{table}

\subsection{Experimental Setup}
All experiments use \texttt{Qwen2.5-VL-7B-Instruct} as the backbone. Post-training proceeds in two stages: (i) SFT on \dataset image--question--answer--reasoning tuples, which establishes a source-grounded reasoning initialization; and (ii) RL with verifiable rewards with GRPO~\cite{shao2024deepseekmath} using a held-out training split of \dataset. Additional details on data splits are provided in Appendix~\ref{app:train_data}, and full training details are provided in Appendix~\ref{app:train_detail}.
For SFT training across all configurations we hold the model architecture, total optimization steps, and learning-rate schedule fixed, so observed differences are attributable to the training signal. We adapt the language layers (attention and MLP
modules) with LoRA at rank $r{=}128$, scaling $\alpha{=}256$, and
dropout~$0$. We use LoRA training due to computational constraints. During both the training stages, the vision tower is frozen. Optimization uses AdamW
(learning rate $1{\times}10^{-4}$, weight decay $0.01$) with a linear
warmup over the first $5\%$ of steps. We train for $3$ epochs at an
effective batch size of $64$ with a maximum sequence length of $8{,}192$
tokens on two NVIDIA H200 GPUs.

For GRPO, we use a mixed medical VQA training set composed of held-out examples from \dataset{} together with four established medical VQA benchmarks (5,120). Table~\ref{tab:grpo_training_data} summarizes the number of examples sampled from each source. All GRPO examples are drawn from held-out splits that were not used during SFT.
To evaluate the impact of \dataset, we benchmark our trained SFT- and RL-tuned models on seven VQA benchmarks (SLAKE, VQA-RAD, PathVQA, PMC-VQA, MedXpertQA, JAMA, and \dataset-Bench) and seven classification benchmarks (HAM10000, EyePACS, HyperKvasir, BrainTumorMRI, VinDr-CXR, VinDr-Mammo, and BUSI). More details on the evaluation datasets are provided in Appendix~\ref{app:eval_datasets}.

\subsection{Results}

\subsubsection{Impact of \dataset in Medical Reasoning} \label{sec:rq1}
\textbf{Improved performance at each post-training stage.} 
The results presented in Table \ref{tab:increment} show a consistent improvement over the data-ablated variants. We observe that training on \dataset samples from existing VQA datasets improves in-distribution performance ($60.47 \rightarrow 66.39$), while incorporating samples corresponding to OpenPMC significantly boosts out-of-distribution performance ($28.98 \rightarrow 32.29$). Moreover, applying GRPO on top of the SFT checkpoint further improves overall performance, yielding a cumulative gain of \gainbase~over the base model, which suggests that outcome-level reward helps refine the policy effectively.

\begin{table*}[t]
\centering
\caption{Comparison of medical vision-language models across VQA and classification benchmarks. Training scale is summarized for reference. \textbf{Bold} = best, \underline{underline} = second-best (among non-gray rows). Gray rows denote large-scale models shown for reference only.
${^*}$ denotes Qwen2.5-VL-7B further trained with SFT and GRPO on \dataset.
}
\label{tab:main_results}
\setlength{\tabcolsep}{5pt}
\small

\resizebox{\textwidth}{!}{%
\begin{tabular}{lc|ccccccc|c}
\toprule
&&\multicolumn{7}{c}{\textbf{VQA Benchmarks}} &\\
\midrule
\textbf{Model} & \textbf{Medical Samples}
& \textbf{\dataset-Bench} & \textbf{SLAKE} & \textbf{VQA-Rad} & \textbf{PathVQA}
& \textbf{PMC} & \textbf{MedX} & \textbf{JAMA} & \textbf{Avg} \\
\midrule
{Lingshu 7B}    & {5M} & {63.20} & {74.20} & {71.23} & {71.96} & {58.75} & {23.51} & {38.95} & {57.40} \\
{OctoMed 7B}    & {8M} & {51.16} & {78.37} & {74.20} & {58.90} & {57.14} & {33.32} & {41.04} & {56.30} \\
{MedGemma 27B}  & {30M} & {48.22}    & {76.20} & {68.12} & {67.20} & {45.50} & {33.70} & {33.25}    & {53.17} \\
MedGemma 1.5 4B   & 30M & 46.19 & 71.39 & 68.12 & 49.58 & 45.39 & \textbf{27.79} & \textbf{41.48} & 49.99 \\
QoQ-Med-VL 7B     & 2.6M & 47.86 & \underline{75.85} & \textbf{73.21} & \underline{64.12} & 51.39 & 21.40 & 38.09 & 53.13 \\
MedVL-Thinker 7B  & 200k & \underline{52.03} & 75.00 & 70.02 & 63.10 & \underline{53.31} & \underline{23.80} & 38.20 & \underline{53.64} \\ 

\multicolumn{10}{l}{\footnotesize \textit{The above numbers are presented for reference only, a direct comparison may not be appropriate due large variations in model size and training samples.}}\\
\midrule
Qwen2.5-VL-7B        & N/A & 47.11 & 62.26 & 67.13 & 63.88 & 49.60 & 22.62 & 35.34 & 49.71 \\
\rowcolor{green!10} ~~~+~\dataset (Ours)${^*}$   & 450K  & \textbf{78.51} & \textbf{85.10} & \underline{72.51} & \textbf{64.10} & \textbf{55.20} & \underline{24.95} & \underline{39.97} & \textbf{60.04} \\
\bottomrule
\end{tabular}%
}

\vspace{8pt}

\resizebox{\textwidth}{!}{%
\begin{tabular}{lc|ccccccc|c}
\toprule
&&\multicolumn{7}{c}{\textbf{Classification Benchmarks}} &\\
\midrule
\textbf{Model} & \textbf{Medical Samples}
& \textbf{HAM} & \textbf{EyePACS} & \textbf{Kvasir} & \textbf{BrainMRI}
& \textbf{VinDr-CXR} & \textbf{VinDr-Mammo} & \textbf{BUSI} & \textbf{Avg} \\
& & \scriptsize Derm.\ & \scriptsize Retina & \scriptsize Endoscopy & \scriptsize Brain MRI
& \scriptsize Chest X-ray & \scriptsize Mammography & \scriptsize Ultrasound & \\
\midrule
{Lingshu 7B}    & {5M} & {21.87} & {59.80} & {64.76} & {78.11} & {71.53} & {61.60} & {74.64} & {61.76} \\
{OctoMed 7B}    & {8M} & {38.14} & {55.40} & {62.71} & {71.42} & {55.84} & {33.20} & {66.03} & {54.68} \\
{MedGemma 27B}  & {30M} & {39.90} & {51.80} & {54.73} & {40.55} & {72.63} & {56.40} & {51.92} & {52.56} \\
MedGemma 1.5 4B   & 30M & 32.92 & \underline{59.70} & 47.61 & 41.67 & 68.63 & \underline{43.00} & 48.72 & 48.89 \\
QoQ-Med-VL 7B     & 2.6M & \textbf{60.11} & 51.30 & \underline{55.48} & \underline{62.22} & \underline{68.98} & 22.00 & \textbf{64.10} & \underline{54.88} \\
MedVL-Thinker 7B  & 200K & 17.62 & 49.80 & 52.44 & 30.91 & 52.17 & 35.60 & 56.69 & 42.18 \\ 
\multicolumn{10}{l}{\footnotesize \textit{The above numbers are presented for reference only, a direct comparison may not be appropriate due large variations in model size and training samples.}}\\
\midrule
Qwen2.5-VL-7B        & N/A & 19.82 & 52.10 & 46.33 & 36.00 & 62.69 & 29.10 & 55.77 & 43.12 \\
\rowcolor{green!10} ~~~+~\dataset (Ours)${^*}$   & 450K  & \underline{58.71} & \textbf{59.50} & \textbf{71.60} & \textbf{75.35} & \textbf{70.07} & \textbf{60.00} & \underline{60.26} & \textbf{65.07} \\
\bottomrule
\end{tabular}%
}
\end{table*}


\textbf{Comparison against medical LVLMs.}
We benchmark our final checkpoint against diverse medical LVLMs (Table~\ref{tab:main_results}), including large-scale medical LVLMs trained on substantially larger medical corpora and recent medical LVLMs at comparable scale and budget.
On VQA, our model achieves the strongest average performance among comparable-scale LVLMs, with a relative gain of $11.7\%$ over the next-best comparable model (MedVL-Thinker~7B, $53.64 \rightarrow
59.94$). More importantly, it remains competitive with the large-scale tier: it surpasses MedGemma~27B by $12.7\%$ relative on average and trails the best large-scale model (Lingshu~7B) by $4.2\%$ ($59.94$ vs.\ $57.40$). On classification, the gap is even larger, where \dataset attains a $50.9\%$ relative gain over the base model, $18.6\%$ over the next-best comparable model (QoQ-Med-VL~7B, $54.88$), and $23.8\%$ over MedGemma~27B. It also surpasses every large-scale reference model on classification, including Lingshu~7B ($+5.4\%$ relative; $61.76 \rightarrow 65.07$).

\subsubsection{Diagnostic Evaluation of LVLMs in Medical Reasoning.}
\label{sec:rq2}
\textbf{Capability-decomposed evaluation on \dataset~Bench.}
We use \dataset-Bench (Section~\ref{sec:reasoning-benchmark}) to decompose performance across three capability axes: perception, medical knowledge, and rationale. For each axis we report \emph{presence} (fraction of required claims mentioned in the candidate trace), \emph{correctness} (fraction of mentioned claims described accurately), and the per-axis \emph{score} $=\text{presence}\times\text{correctness}$. Results are summarized in Table~\ref{tab:reasoning_trace_eval}. \dataset attains the highest Reasoning Trace Score, exceeding leading closed-source baselines such as (\texttt{gpt-5-mini}) by $6.9$ points, corresponding to a $9.7\%$ relative gain and the medical LVLM  by $18.0$ points. The largest gain is observed in \emph{presence}: relative to the best baseline on each axis, our model leads presence by $+3.5$ on perception, $+8.5$ on medical knowledge, and $+4.3$ on reasoning. The pattern is most pronounced for medical knowledge, where \dataset surfaces $89.2\%$ of required clinical claims compared with $80.7\%$ for the next-best system. Compared with the same Qwen2.5-VL-7B backbone before our
post-training, presence improves by $+29$ to $+42$ points across the three axes, far exceeding the corresponding correctness gains. We read this
asymmetry as a direct effect of source-grounded supervision: training on
figure-derived traces teaches the model \emph{which} visual and contextual
cues are relevant to a given answer. Table~\ref{tab:reasoning_unit_highlight} shows a detailed example of capability decomposition.








\newcolumntype{Y}{>{\centering\arraybackslash}X}


\begin{table*}[t]
\centering
\footnotesize
\setlength{\tabcolsep}{5pt}
\renewcommand{\arraystretch}{1}
\caption{\textbf{Capability-decomposed reasoning trace evaluation on \dataset-Bench.} For each capability axis (Perception, Medical Knowledge,
Rationale), \textit{Presence} (Pres.) is the fraction of reference claims
surfaced in the candidate trace, \textit{Correctness} (Corr.) is the fraction
of those mentioned claims that are accurate, and the per-axis \textit{Score}
is their product ($\text{Pres.}\times\text{Corr.}$). The \textbf{Reasoning
Trace Score} is the unweighted mean of the three per-axis scores. Top block:
closed-source frontier models. Middle block: open-weight 7B baselines
(general and medical). Bottom: our model. Best result per column in
\textbf{bold}; second-best \underline{underlined}.
}
\label{tab:reasoning_trace_eval}
\begin{tabularx}{\textwidth}{@{}l *{9}{Y} c c}
\toprule
\multirow{2}{*}{\textbf{Model}}
& \multicolumn{3}{c}{\textbf{Perception}}
& \multicolumn{3}{c}{\textbf{Medical Knowledge}}
& \multicolumn{3}{c}{\textbf{Rationale}}
& \multirow{2}{*}{\textbf{Trace}} 
& \multirow{2}{*}{\textbf{Answer}} \\
\cmidrule(lr){2-4} \cmidrule(lr){5-7} \cmidrule(lr){8-10}
& Pres. & Corr. & Score & Pres. & Corr. & Score & Pres. & Corr. & Score & Score & Acc. \\
\midrule
\texttt{claude-haiku-4.5}
  & 62.8 & 84.2 & 52.9
  & \underline{80.7} & 93.0 & \underline{75.1}
  & 79.9 & 92.0 & 73.5
  & 67.1 & 57.3\\
\texttt{gpt-5-mini}
  & 67.9 & \underline{91.1} & 61.9
  & 76.4 & 92.6 & 70.8
  & \underline{83.6} & \underline{95.9} & \underline{80.2}
  & \underline{70.9} & 67.7\\
\texttt{gemini-3-flash}
  & \underline{70.8} & \textbf{91.2} & \underline{64.6}
  & 73.8 & \underline{94.3} & 69.6
  & 80.4 & \textbf{96.1} & 77.3
  & 70.5 & \underline{72.0}\\
OctoMed 7B
  & 58.4 & 81.2 & 47.4
  & 76.1 & 89.8 & 68.3
  & 72.7 & 87.4 & 63.6
  & 59.8 & 51.2 \\ \midrule
Qwen2.5-VL 7B
  & 45.1 & 76.0 & 34.3
  & 47.2 & 73.0 & 34.5
  & 58.9 & 73.8 & 43.5
  & 37.4 & 47.1 \\
\textbf{~~~+~\dataset (Ours)}
  & \textbf{74.3} & 87.8 & \textbf{65.2}
  & \textbf{89.2} & \textbf{95.3} & \textbf{85.0}
  & \textbf{87.9} & 94.5 & \textbf{83.1}
  & \textbf{77.8} & \textbf{78.5}\\
\bottomrule
\end{tabularx}
\end{table*}

\begin{figure*}[t]
    \centering
    \includegraphics[width=\textwidth]{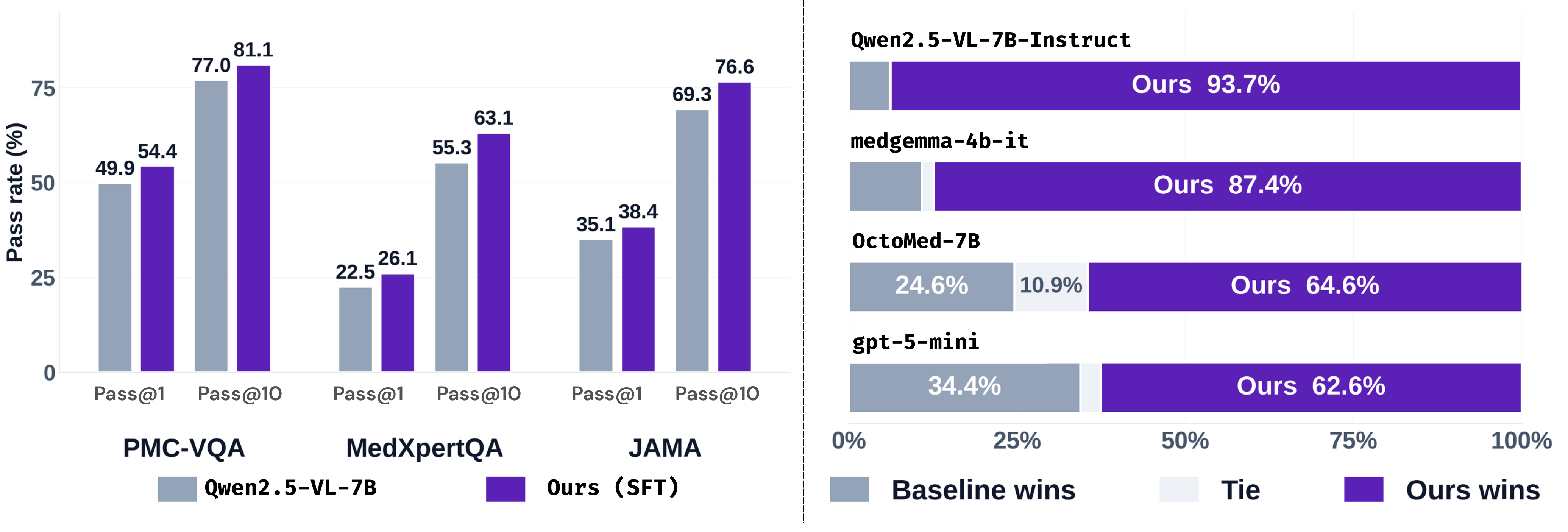}
    \caption{\textbf{Performance of our model.}
    \textbf{(Left) Pairwise win / tie / lose rates} from our model's perspective against four baselines, judged head-to-head on medical VQA traces. \textit{Baseline wins} (slate) denotes the baseline trace was preferred, \textit{Tie} (light gray) indicates the judge marked the responses as equivalent, and \textit{Ours wins} (purple) denotes our model was preferred. Our checkpoint (SFT + GRPO) is favored across all four matchups, with the largest margin against the Qwen-2.5-VL backbone before post-training.
    \textbf{(Right) Pass@1 and Pass@10} of the SFT checkpoint compared to the Qwen-2.5-VL backbone across three medical VQA benchmarks (PMC-VQA, MedXpertQA, JAMA). Our model improves over the backbone at both $k\!=\!1$ and $k\!=\!10$ on all three datasets.}
    \label{fig:SFT-plot}
\end{figure*}

\textbf{Reasoning-trace quality and sampling reliability.}
We probe both: pairwise preferences, in which the judge model (\texttt{gpt-5-mini}) selects the better trace from each pair on $N{=}1500$ shared instances (Figure~\ref{fig:SFT-plot}-Left); and Pass@$k$, probability of obtaining at least one correct answer among $k$ sampled responses, providing a proxy for the model's reasoning reliability under stochastic decoding (Figure~\ref{fig:SFT-plot}-Right). Our model is preferred in a majority of comparisons against five of the six baselines. The largest margin is against the same Qwen2.5-VL-7B backbone before our post-training. Margins remain large against medical baselines. These results suggest that judges favor traces that surface more of the relevant clinical and visual evidence, even when individual-claim correctness is comparable to that of the strongest closed-source systems. Across all three benchmarks, our model improves over the backbone at both $k{=}1$ and $k{=}10$. On MedXpertQA and JAMA, the two
benchmarks furthest from the SFT distribution, the relative gain at $k{=}10$ exceeds the gain at $k{=}1$, indicating that SFT on \dataset expands its reachable-solution capacity: correct reasoning paths that were absent from the backbone's output distribution become accessible
after training, even on clinically demanding out-of-distribution problems.

\section{Expert Review and Case Analysis}
A board-certified internist independently reviewed a random sample of $100$ cases through a custom
annotation interface (Figure~\ref{fig:expert} in Appendix~\ref{app:case-review}), assigning binary judgements along five quality dimensions. The marked answer was confirmed as correct in $100/100$ cases; the generated reasoning was judged faithful to the image context in $100/100$; the
question was rated clinically meaningful in $100/100$ and answerable from the evidence shown in $100/100$; and the imaging-modality label was verified in $99/100$. These results indicate that \dataset--Bench items satisfy expert criteria for answer correctness, evidential grounding, and clinical relevance. 
Broader implications and responsible-use considerations are discussed in the Impact Statement (Appendix~\ref{app:impact}).


\section{Conclusion}

\noindent\textbf{Summary.} 
We introduced \dataset, a source-grounded VQA dataset of medical reasoning traces distilled from scientific articles together with \dataset~Bench, a paired benchmark that decomposes trace quality across perception, medical knowledge, and reasoning. Post-training a 7B backbone on \dataset first by SFT, then by GRPO yields a \gainbase~relative gain in average accuracy over the base model and produces traces preferred over five of six baselines, including frontier closed-source systems that are substantially larger. The capability
decomposition isolates where this improvement comes from: source-grounded
supervision primarily improves the model's ability to \emph{surface} the
visual and contextual evidence relevant to a clinical question, while GRPO adds complementary gains through outcome-level reward. A clinician audit indicates that the benchmark closely tracks expert judgment.

\noindent\textbf{Limitation.} 
\dataset is derived from published biomedical figures and therefore inherits the biases of the source literature, including over-representation of rare or pedagogically salient cases and uneven modality coverage.
Although the dataset uses multi-stage filtering, LLM-based generation, and expert review, its reasoning traces and benchmark scores should not be interpreted as evidence of clinical safety or readiness for deployment.

\noindent\textbf{Broader Impact and Societal Considerations}
\dataset is designed to enable reproducible research on medical LVLM reasoning. Two design choices are intentionally pro-social. First, every reasoning step is anchored to a specific figure and its case-level context, so downstream errors can be traced to concrete evidence rather than an opaque model state. Second, because the dataset, benchmark, and trained checkpoint are open and run on a single 7B backbone, academic and clinical groups can study where their models attend and where they fail without depending on frontier closed-source systems.

 
\clearpage
{
\small
\bibliographystyle{unsrt}
\bibliography{main}
}

\clearpage
\appendix

\section{Data Curation Details}
\label{app:datacuration}
\subsection{Multi-Level Quality Filtering}
\label{app:filtering}
\subsubsection{Visual Quality} 
\label{app:visualfilter}

\begin{figure}[t]
    \centering
    \includegraphics[width=\linewidth]{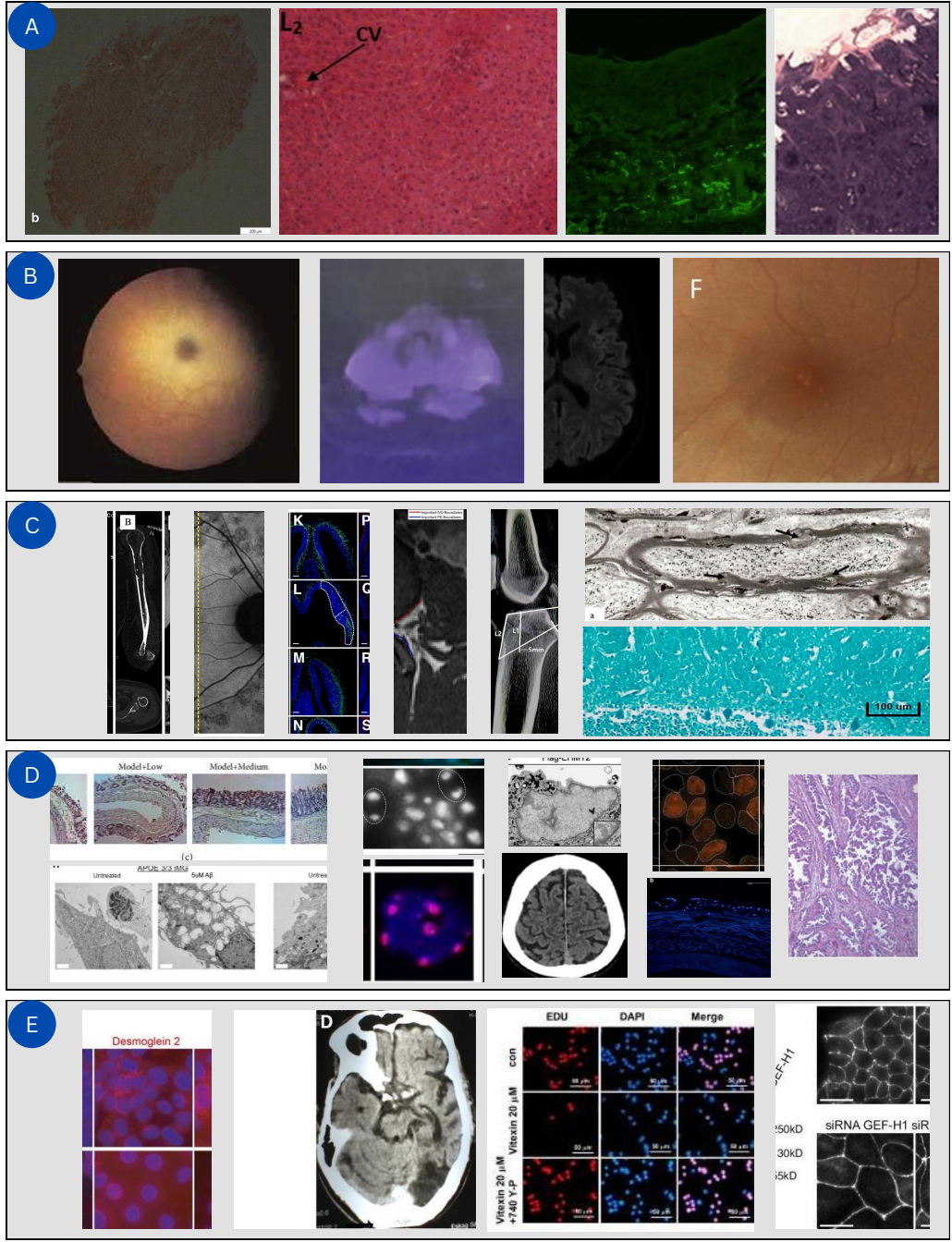}
    \caption{
    Examples of images removed by the visual quality filtering stage.
    The filtering criteria exclude images with (A) blurry visual content, (B) visible compression artifacts, (C) extreme aspect ratios, (D) too small image size, and (E) excessive white borders.
    These examples illustrate cases that do not provide reliable perceptual grounding for downstream medical visual reasoning.
    }
    \label{fig:visual_quality_filter_examples}
\end{figure}

The visual-usability filter described in Section~\ref{sec:quality-filtering} applies four pixel-level checks; an image is rejected if it fails any of them. Representative discarded examples are shown in Figure~\ref{fig:visual_quality_filter_examples}.

\textbf{Spatial resolution.} Shorter side $\geq 224$ pixels.

\textbf{Layout validity.} Aspect ratio $\leq 3{:}1$, and the outer $15\%$ of the image contains less than $35\%$ near-white pixels. The border check is a heuristic for slide-style panels, heavily padded figures, and other presentation artifacts.

\textbf{Clarity.} Laplacian variance $\geq 60$ (a sharpness floor) and an estimated effective resolution of at least $224 \times 224$, to exclude up-sampled or otherwise low-detail images.

\textbf{Artifact burden.} We discard images with visible compression artifacts (e.g., $8{\times}8$ blockiness from heavy JPEG quantization) or other degradations that obscure clinically relevant content.

\subsubsection{Textual Filtering}
\label{app:textfiltering}

The text-and-context filter (Section~\ref{sec:quality-filtering}) is applied to each image--context pair that passes the visual-usability stage. The classifier returns a binary keep/reject decision; a pair is kept only when all four criteria (textual adequacy, clinical relevance, image--text alignment, reasoning readiness) are satisfied.

\textbf{Classifier.} We use \texttt{<MODEL>} as a zero-shot judge.\\
\textbf{Prompt.} The prompt presents the four criteria as an explicit rubric and asks the model to issue a per-criterion verdict followed by a final keep/reject label and a one-sentence justification. The full template, including the rubric definitions and the output schema, is shown in Prompt~\ref{prompt:text_quality_filter}. The prompt's six PASS conditions map onto the four main-paper criteria as follows: textual adequacy aggregates conditions~1 and~4; clinical relevance aggregates conditions~2 and~3; image--text alignment corresponds to condition~5 (used only when the image is provided to the judge); and reasoning readiness corresponds to condition~6.\\
\textbf{Decision rules.} Beyond the four rubric criteria, we also require the textual context to be in English; non-English captions are rejected at this stage. The most common rejection patterns are: (a) captions that only name a condition or treatment without describing the specific subfigure (textual adequacy); (b) figures depicting schematics, plots, instruments, or animal/bench-only experiments rather than clinically meaningful human cases (clinical relevance); (c) in-text references that describe a different figure or only report cohort-level results (image--text alignment); and (d) contexts too generic to support a defensible question--answer pair (reasoning readiness). Representative rejected examples covering each of these patterns are shown in Table~\ref{tab:text_quality_rejected_examples}, and matching accepted examples in Table~\ref{tab:text_quality_accepted_examples}.\\

\begin{table*}[t]
\centering
\caption{
Diverse examples of image--text pairs rejected during text-quality filtering.
Rejected samples fail for different reasons, including poor image-grounding, insufficient reasoning signal, non-English text, non-human or bench-only content, and non-clinical domains.
}
\label{tab:text_quality_rejected_examples}

\setlength{\tabcolsep}{3pt}
\renewcommand{\arraystretch}{1.18}
\scriptsize

\resizebox{\textwidth}{!}{
\begin{tabular}{p{0.12\textwidth} p{0.14\textwidth} p{0.14\textwidth} p{0.14\textwidth} p{0.14\textwidth} p{0.14\textwidth} p{0.14\textwidth}}
\toprule
 & \textbf{A} & \textbf{B} & \textbf{C} & \textbf{D} & \textbf{E} & \textbf{F} \\
\midrule

\textbf{Failure type} &
Context mismatch &
Non-English text &
Non-human / animal case &
Bench-only animal model &
Non-clinical domain &
Method-only / low usefulness \\

\midrule

\textbf{Sub-caption} &
A pH image obtained during exercise showing different pH values within ROIs of gastrocnemius medialis, gastrocnemius lateralis, and soleus. &
Lésions cutanées vésiculo-bulleuses à contenu hémorragique de taille millimétrique à 1 cm de regroupement herpétiforme. &
Detail on the peduncle and tail of the striped dolphin with the FAD's parts. &
Example 20X image taken from a lung tissue section from a Cdh5-CreERT2; ROSA26LSL-tdTomato; Atf3lox/lox mouse. &
EDX map for the element F of a LiNi$_{0.5}$Mn$_{1.5}$O$_4$ particle cycled in 1 M-LPF-DMDOHD electrolyte. &
Retinal fundus image assessed using VAMPIRE software. Arterioles, venules, and deleted segments are indicated. \\

\midrule

\textbf{Context summary} &
With the advent of 7T MR scanners, it is important to clarify the benefits of 7T compared with 3T, including scan-time reduction, increased spatial resolution, SNR, and CNR. The discussion focuses on MRI acquisition properties and protocol-level image quality. &
L'examen cutané a trouvé des lésions cutanées vésiculo-bulleuses à contenu hémorragique, de taille millimétrique à 1 cm, avec un regroupement herpétiforme. The description is clinically meaningful but written in French rather than English. &
The rope was tightly wrapped around the end of the dolphin's peduncle and tail, producing visible necrosis in the surrounding tissues. The case describes an entanglement injury in a striped dolphin rather than a human clinical case. &
Endothelial-specific Atf3 deletion is studied in a mouse model of acute lung injury and tissue regeneration. Lung tissue sections are imaged to evaluate experimental cellular or molecular changes in a preclinical animal model. &
The CEI formed in the 1 M-LPF-DMDOHD electrolyte contains a higher amount of --CF$_3$ and Li$_x$PO$_y$F$_z$ components compared with baseline electrolytes. Elemental mapping is used to characterize fluorine distribution on a cycled battery particle. &
Retinal microvascular parameters, including CRAE, CRVE, AVR, fractal dimension, and tortuosity, are measured from optic-disc-centered fundus images using semi-automated VAMPIRE software. Arterioles, venules, and excluded vessel segments are marked during this measurement process. \\

\midrule

\textbf{Why rejected} &
The context focuses on scanner and protocol properties rather than interpreting the pH differences visible in the image. It does not support a biological or clinical reasoning question grounded in the visual pH map. &
The sample is removed because the textual supervision is not in English, even though the visual content may be clinically relevant. Keeping it would introduce language inconsistency into the English medical VQA dataset. &
The image contains pathology-like visual evidence, but the subject is a dolphin rather than a human patient. It is therefore outside the intended human clinical medical-image QA scope. &
The sample is biomedical but preclinical and bench-oriented. It is less useful for downstream human medical-image QA because the reasoning target concerns an experimental mouse model rather than clinical interpretation. &
The content is materials science rather than biomedical or clinical imaging, making it outside the scope of clinically grounded medical VQA supervision. &
The text is centered on a measurement pipeline rather than a specific clinical finding or abnormality visible in the fundus image. It is method-focused instead of reasoning-focused. \\

\bottomrule
\end{tabular}
}

\end{table*}

\begin{table*}[t]
\centering
\caption{
Examples of image--text pairs accepted by the text-quality filtering stage.
Accepted samples contain sub-caption and image context that are aligned with the visual evidence and support clinically or biologically meaningful reasoning.
}
\label{tab:text_quality_accepted_examples}

\setlength{\tabcolsep}{3pt}
\renewcommand{\arraystretch}{1.18}
\scriptsize

\resizebox{\textwidth}{!}{
\begin{tabular}{p{0.12\textwidth} p{0.14\textwidth} p{0.14\textwidth} p{0.14\textwidth} p{0.14\textwidth} p{0.14\textwidth} p{0.14\textwidth}}
\toprule
 & \textbf{A} & \textbf{B} & \textbf{C} & \textbf{D} & \textbf{E} & \textbf{F} \\
\midrule

\textbf{Reason accepted} &
Image-grounded injury finding &
Clear clinical intervention and outcome &
Staining supports infection localization &
Experimental signal linked to tumor region &
Disease-specific visual morphology &
Cell-type and localization reasoning \\

\midrule

\textbf{Sub-caption / image text} &
Extensive leakage of the 10 kDa dextran rhodamine B is observed around sites of injury at 24 h post-injury. &
3D angiographic reconstructions from left ICA post clipping and coiling with the giant aneurysm completely obliterated by the microsurgical clip and coil mass. &
Brain sections were stained with anti-NP antibody and DAPI. Sections were obtained from uninfected and infected snakes. &
RFP-labeled hydrogels and GFP-labeled tumor cells in the tumor area at 2 weeks after IPN hydrogel treatment were detected by fluorescence microscope. &
Median and ulnar nerve in CIDP, anti-MAG neuropathy, d-CIDP, and healthy subject. &
Co-labeling of P7 cerebellum with antibodies against GFP and S100$\beta$. A typical staining pattern for radial glia is seen. \\

\midrule

\textbf{Context summary} &
Different-size permeability tracers exhibit distinct temporal patterns of blood--spinal-cord barrier dysfunction after spinal cord injury. At 24 h post-injury, smaller tracers up to 10 kDa are observed outside blood vessels in and around the injury center, while larger tracers remain more confined to intact vessels. &
A residual giant carotid ophthalmic aneurysm is treated by coiling following neck reconstruction after clipping. Angiographic reconstructions show the aneurysm before and after treatment, including the clip-reconstructed neck and coil mass, with complete post-treatment obliteration. &
NP staining is detected in brain tissue of infected ball pythons but not in most other examined tissues. Infected and uninfected samples are compared using anti-NP antibody and DAPI staining, linking the visual signal to infection status and tissue localization. &
After IPN hydrogel treatment and portal-vein embolization in an orthotopic HCC model, fluorescence microscopy shows RFP-labeled hydrogel signals together with GFP-labeled tumor cells in the tumor area. The visual signal links treatment localization with tumor-region tissue effects. &
Very-high-resolution peripheral nerve ultrasound is used to examine median and ulnar nerves across CIDP, anti-MAG neuropathy, d-CIDP, and healthy control. Nerve cross-sectional area and fascicle morphology are compared across disease groups. &
P2X7 localization is studied using EGFP-specific staining in brain slices from multiple mouse lines. In the P7 cerebellum, clearer GFP signal enables comparison with S100$\beta$-positive radial or Bergmann glia morphology. \\

\midrule

\textbf{Why accepted} &
The visual leakage pattern is directly linked to barrier disruption, tracer size, and a post-trauma window for delivery of small therapeutic compounds. &
The image and text are aligned around a visually verifiable treatment outcome: complete aneurysm obliteration after clipping and coiling. &
The image context directly supports reasoning about infection status, tissue localization, and differences between infected and uninfected samples. &
The image context connects treatment, hydrogel localization, tumor-cell signal, and tumor-region tissue effects. &
The image supports comparison of disease-specific nerve morphology across pathological and healthy cases. &
The image supports reasoning about cell type, staining pattern, and P2X7 localization in cerebellar tissue. \\

\bottomrule
\end{tabular}
}

\end{table*}

\clearpage

\begin{promptbox}{Text-Quality and Reasoning-Signal Filtering Prompt}
\label{prompt:text_quality_filter}
You are a Senior Medical Content Analyst and strict data-quality gatekeeper for a medical Visual Question Answering (VQA) pipeline.

\vspace{0.4em}
\textbf{Task.}
Given one row containing text associated with a medical sub-figure, and optionally the image itself, decide whether it should \textbf{PASS} or \textbf{FAIL}.

\vspace{0.4em}
\textbf{PASS} only if all are true:
\begin{enumerate}
    \item The text has acceptable English quality.
    \item The content is human/clinical and relevant to patient-level medical interpretation.
    \item The content is not primarily non-human, veterinary, animal-model, bench-only, or non-medical.
    \item The text is informative enough for downstream medical-image QA.
    \item If an image is provided, it is usable for visual medical QA.
    \item The text contains reasoning signals that support a complex image-grounded QA pair.
\end{enumerate}

\vspace{0.4em}
\textbf{Reasoning signals include:}
\begin{itemize}
    \item \textbf{Causal:} explains the cause or effect of a visual feature.
    \item \textbf{Comparative:} contrasts the sub-figure with another condition, baseline, or group.
    \item \textbf{Methodological/Functional:} explains how a mechanism works or why a visual pattern appears.
\end{itemize}

\vspace{0.4em}
\textbf{FAIL} if the text is purely navigational, tautological, low-information, not grounded in the specific sub-figure, non-human/non-clinical, or paired with an unusable image.

Prioritize high precision. Keep only rows where the text supports a question requiring both visual inspection and textual knowledge.

\vspace{0.4em}
\textbf{Return only valid JSON:}
\begin{verbatim}
{
  "decision": "PASS" or "FAIL",
  "reasons": ["short reason 1", "short reason 2"]
}
\end{verbatim}

Do not include extra keys or extra text.

\end{promptbox}

\clearpage

\begin{table*}[t]
...
\end{table*}

\subsection{Clinical Task Taxonomy}
\label{app:clinicaltax}
The clinical-task taxonomy presented in Table~\ref{tab:task-taxonomy} was developed through an inductive, bottom-up analysis of a randomly sampled subset of the dataset. Two annotators independently reviewed the sampled questions and iteratively grouped them according to their underlying clinical intent. Semantically overlapping groupings were progressively merged into higher-level themes. The resulting categories were then organized into five task families that mirror the canonical stages of clinical reasoning (Perception, Diagnosis, Management, Risk, and Workup) yielding the final taxonomy used for question generation.

Later, we use the resulting clinical task taxonomy to assign each image--context pair to all applicable task categories. A pair is considered eligible for a category only when its associated context contains sufficient information to support the generation of a grounded question and answer for that task, including enough explanatory evidence to justify why the answer is correct. This constraint ensures that the selected categories can later support reasoning-trace construction rather than only final-answer generation. The prompt used for this category-assignment step is shown in Prompt~\ref{prompt:question-category-assignment}.

\begin{table}[t]
\centering
\footnotesize
\setlength{\tabcolsep}{6pt}
\renewcommand{\arraystretch}{1.1}
\caption{Clinical-task taxonomy used for question generation, organized into five task families: Perception (visual description and localization), Diagnosis (diagnostic reasoning), Management (treatment and intervention), Risk (outcomes, safety, complications), and Workup (follow-up evaluation).}
\label{tab:task-taxonomy}
\begin{tabularx}{\linewidth}{@{} l l X @{}}
\toprule
\textbf{Family} & \textbf{Category} & \textbf{Question objective} \\
\midrule
\multirow{6}{*}{\textit{Perception}}
  & Findings / description      & Identify the main visible finding or visual pattern. \\
  & Anatomy / localization      & Identify the depicted structure, tissue, or organ. \\
  & Normal vs.\ abnormal        & Decide whether the appearance is normal or abnormal. \\
  & Annotation / marker         & Annotate a visual marker (arrow, box, label, etc.). \\
  & Spatial location            & Locate a finding within the image or anatomy. \\
  & Counting                    & Count visible structures, lesions, or cells. \\
\cmidrule(lr){2-3}
\multirow{4}{*}{\textit{Diagnosis}}
  & Diagnosis                   & Infer the most likely diagnosis from visual and clinical cues. \\
  & Mechanism / pathophysiology & Explain the mechanism underlying the finding. \\
  & Differential diagnosis      & Distinguish the correct diagnosis from plausible alternatives. \\
  & Severity grading            & Assess the grade, stage, or extent of the condition. \\
\cmidrule(lr){2-3}
\multirow{3}{*}{\textit{Management}}
  & Next-step management        & Choose the appropriate management or intervention. \\
  & Surgical management         & Reason about a surgical approach or intraoperative decision. \\
  & Drug therapy                & Identify the relevant medication or therapeutic class. \\
\cmidrule(lr){2-3}
\multirow{5}{*}{\textit{Risk}}
  & Complications               & Identify an associated complication or adverse event. \\
  & Prognosis                   & Infer the expected outcome or disease course. \\
  & Symptoms / signs            & Link the finding to associated symptoms or signs. \\
  & Safety / contraindications  & Flag safety concerns or contraindications. \\
  & Hereditary risk             & Reason about inherited risk or genetic association. \\
\cmidrule(lr){2-3}
\textit{Workup} & Next-step diagnostic test & Choose the appropriate follow-up test or evaluation. \\
\bottomrule
\end{tabularx}
\end{table}

\begin{promptbox}[breakable]{Question Category Assignment Prompt}
\label{prompt:question-category-assignment}

\textbf{Role.}

You are a medical visual question answering dataset curator. Your task is to analyze an image--context pair and assign the question categories that are sufficiently supported by the provided context.

\vspace{0.5em}
\noindent\textbf{Task.}

Given the modality and textual context associated with a medical image, select all applicable question categories that can be used to generate grounded questions from the context.

\vspace{0.5em}
\noindent\textbf{Input.}

The prompt is provided with the following fields:

\begin{itemize}
    \item \textbf{Modality:} the imaging or visual modality of the example.
    \item \textbf{Image Context:} the textual context associated with the image.
\end{itemize}

\vspace{0.5em}
\noindent\textbf{Category selection rules.}
\begin{itemize}
    \item Select all categories that apply to the image--context pair.
    \item A category is eligible only if the provided context contains enough explicit explanation or supporting evidence to generate a grounded question and answer for that category, including sufficient rationale for why the answer is correct so that a reasoning trace can be derived from the context.
    \item Do not select categories that are only weakly implied or unsupported by the context.
    \item The selected category name must exactly match one of the allowed category names.
\end{itemize}

\vspace{0.5em}
\noindent\textbf{Allowed category names.}

The text content inside each \verb|<category>...</category>| element must exactly match one of the following category names:

\begin{itemize}
    \item \textbf{Diagnosis}: [description]
    \item \textbf{Differential diagnosis}: [description]
    \item \textbf{Next-step diagnostic test or imaging}: [description]
    \item \textbf{Next-step treatment / management}: [description]
    \item \textbf{Surgery / operative management}: [description]
    \item \textbf{Drug therapy / pharmacologic treatment}: [description]
    \item \textbf{Safety / contraindications and adverse effects}: [description]
    \item \textbf{Findings / description only}: [description]
    \item \textbf{Prognosis / risk assessment}: [description]
    \item \textbf{Future risk / hereditary probability}: [description]
    \item \textbf{Complication or adverse event}: [description]
    \item \textbf{Anatomy / localization}: [description]
    \item \textbf{Spatial location on image (quadrant / region)}: [description]
    \item \textbf{Normal vs abnormal}: [description]
    \item \textbf{Severity grading}: [description]
    \item \textbf{Counting}: [description]
    \item \textbf{Symptom}: [description]
    \item \textbf{Annotation / marker interpretation}: [description]
    \item \textbf{Mechanism / pathophysiology explanation}: [description]
    \item \textbf{Other clinical reasoning}: [description]
\end{itemize}

\vspace{0.5em}
\noindent\textbf{Output format.}

Return only the following strict XML format:

\begin{verbatim}
<question_categories>
  <category>Diagnosis</category>
  <category>Next-step treatment / management</category>
  ...
</question_categories>
\end{verbatim}

Do not include explanations, comments, markdown, or any text outside the XML block.

\end{promptbox}
\clearpage

\begin{figure*}[t]
    \centering
    \includegraphics[width=\textwidth]{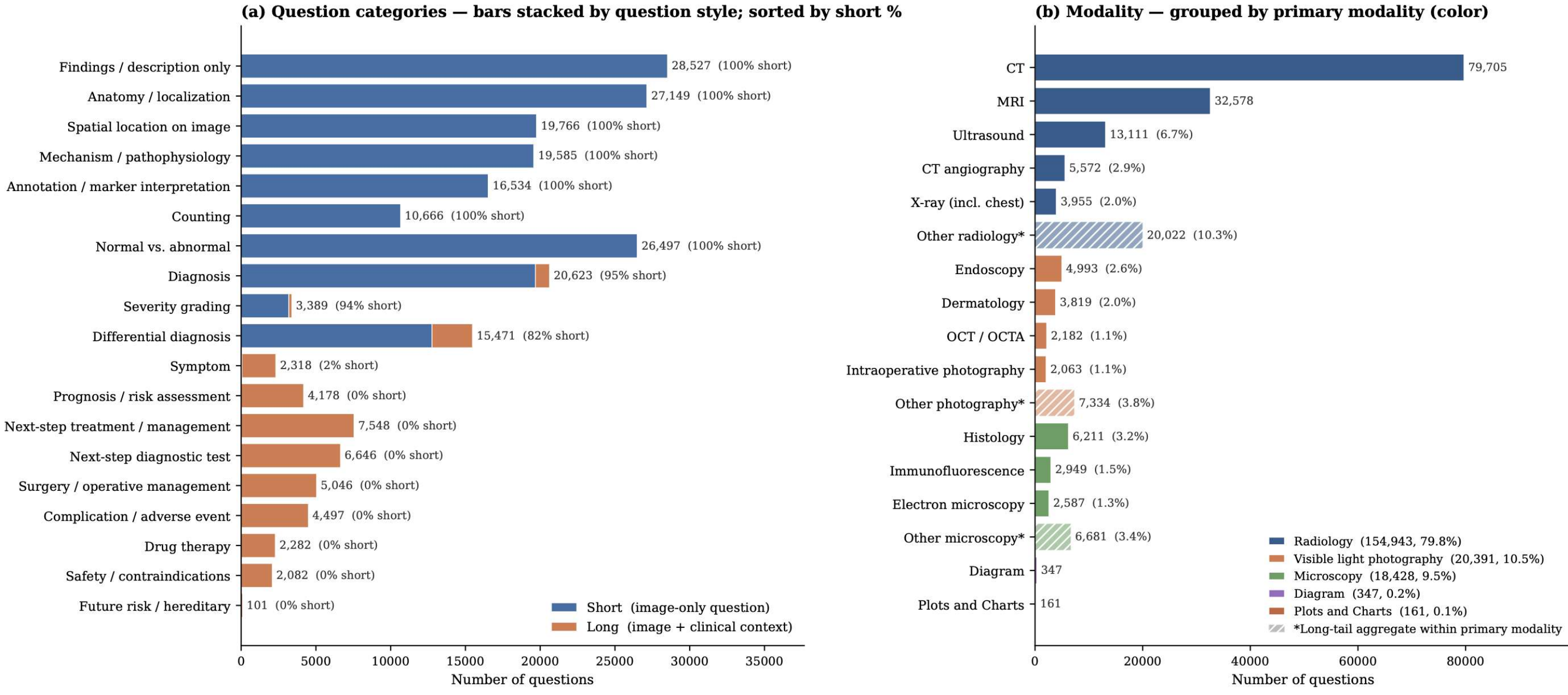}
    \caption{
    Dataset statistics for \dataset. 
    (A) Distribution of question categories, with bars stacked by question style, showing the balance between short image-only questions and long image-plus-clinical-context questions. 
    (B) Distribution of imaging modalities, grouped by primary modality, illustrating the diversity of radiology, visible-light photography, microscopy, diagrams, and plots/charts represented in the dataset.
    }
    \label{fig:benchmark_stats}
\end{figure*}

\subsection{Question Construction}
\label{app:questiongen}
The five validity constraints summarized in Section~\ref{sec:quality-filtering} (source grounding, self-contained answerability, visual necessity, anti-leakage, non-visual long context) are enforced through a single prompt template that wraps modality- and category-specific instructions. Generation is framed as a constrained validity decision; whenever the assigned task cannot be instantiated from the available evidence, the model returns \texttt{INVALID} and the case is dropped. Answers take one of four formats determined by the task category: four- or five-option multiple choice, \textsc{Yes/No}, \textsc{True/False}, or \textsc{Normal/Abnormal}. Some examples of obtained questions for each category are shown in Table~\ref{tab:question_examples}.

\subsubsection{Question Types}
\label{app:QuestionExamples}

Each item is also tagged \emph{short-context} or \emph{long-context}; some categories support both. Short-context stems are image-forward, supplying little or no clinical information beyond the image. Long-context stems supply non-visual clinical detail such as history, symptoms, laboratory findings and treatment course before posing the question. Stem style is matched to the task: perception-heavy categories use short stems to keep the image essential, while decision-oriented categories use long stems to define the clinical problem. Cases whose required context is missing, or whose context would let a text-only solver answer the question, are discarded.

\textbf{USMLE-style item.} Each generated item follows the structure of a board-style clinical question: a short, purposeful vignette (typically specifying patient demographics, relevant history, and key symptoms) followed by a question and a small set of answer options. Every detail in the stem is required for the answer; no extraneous narration is included. A typical opening pattern is \emph{“A 58-year-old woman presents with … for the past 3 weeks …”}, after which the model is asked to infer the answer from the vignette together with the image. Across all categories, items are constrained to standard board-style framing, favoring commonly accepted clinical interpretations over niche distinctions.

\textbf{Model interface.} The generator (\texttt{gpt-5-mini}, \texttt{reasoning\_effort=medium}) receives as input the image, the source context, the target task category, the desired stem style (short or long), and the predefined answer format. It returns a structured JSON object containing the question stem, answer options, the correct-answer label (e.g., A--E), the answer format, the rationale, and the image scope (which sub-figure or panel the question depends on). When any of the five validity constraints cannot be met, the model returns an explicit \texttt{INVALID} marker rather than degrading to a weaker question; \texttt{INVALID} cases are dropped in a downstream filter.

\textbf{Stem-style routing.} Categories are routed to stem styles by their clinical role. Perception-heavy categories use \emph{short} stems to keep the image essential; decision-oriented categories use \emph{long} stems to define the clinical problem; categories that admit either framing receive whichever style the source context can support. The full routing is shown in Table~\ref{tab:stem-routing}.

\begin{table}[h]
\centering
\small
\setlength{\tabcolsep}{6pt}
\renewcommand{\arraystretch}{1.15}
\caption{Routing of clinical-task categories to short and long stem styles.}
\label{tab:stem-routing}
\begin{tabular}{@{}p{0.18\linewidth} p{0.74\linewidth}@{}}
\toprule
\textbf{Stem style} & \textbf{Categories} \\
\midrule
Short only & Findings / description; anatomy / localization; spatial location; counting; annotation / marker; mechanism / pathophysiology. \\
Long only  & Next-step diagnostic test; next-step management; surgical management; drug therapy; safety / contraindications; complications. \\
Either     & Diagnosis; differential diagnosis; prognosis; hereditary risk; symptoms / signs; normal vs.\ abnormal; severity grading. \\
\bottomrule
\end{tabular}
\end{table}

\textbf{Answer-format routing.} The four supported formats---four- or five-option multiple choice, \textsc{Yes/No}, \textsc{True/False}, and \textsc{Normal/Abnormal}---are assigned per category. Multiple choice is the default for diagnostic, management, and workup categories; \textsc{Normal/Abnormal} is restricted to categories whose answer is a binary visual judgment (e.g., normal vs.\ abnormal); \textsc{Yes/No} and \textsc{True/False} are reserved for categories where the source context licenses a binary verification.

\textbf{Category-specific safeguards.} Two categories require additional safeguards beyond the five validity constraints. \emph{Annotation / marker} items are generated only when explicit graphical markers (arrows, circles, boxes, pointers) are visible in the image; panel letters alone do not count. \emph{Spatial localization} items refer to the finding abstractly (``the finding'') and restrict answer options to positional descriptors---upper/lower, left/right, central/peripheral, quadrant, or clock-face---rather than diagnostic terms. These rules prevent diagnostic information from leaking through the stem in two categories where leakage is otherwise hard to detect.
 

\newcolumntype{I}{>{\centering\arraybackslash}m{0.20\textwidth}}
\newcolumntype{C}{>{\raggedright\arraybackslash}m{0.20\textwidth}}
\newcolumntype{Q}{>{\raggedright\arraybackslash}m{0.54\textwidth}}

\onecolumn

\begingroup
\small
\setlength{\tabcolsep}{3pt}
\renewcommand{\arraystretch}{1.08}

\begin{longtable}{I C Q}
\caption{Examples of question categories with corresponding images and questions.}
\label{tab:question_examples} \\

\toprule
\textbf{Figure} & \textbf{Question Category} & \textbf{Question} \\
\midrule
\endfirsthead

\toprule
\textbf{Figure} & \textbf{Question Category} & \textbf{Question} \\
\midrule
\endhead

\midrule
\multicolumn{3}{r}{\textit{Continued on next page}} \\
\endfoot

\bottomrule
\endlastfoot

\includegraphics[width=0.95\linewidth]{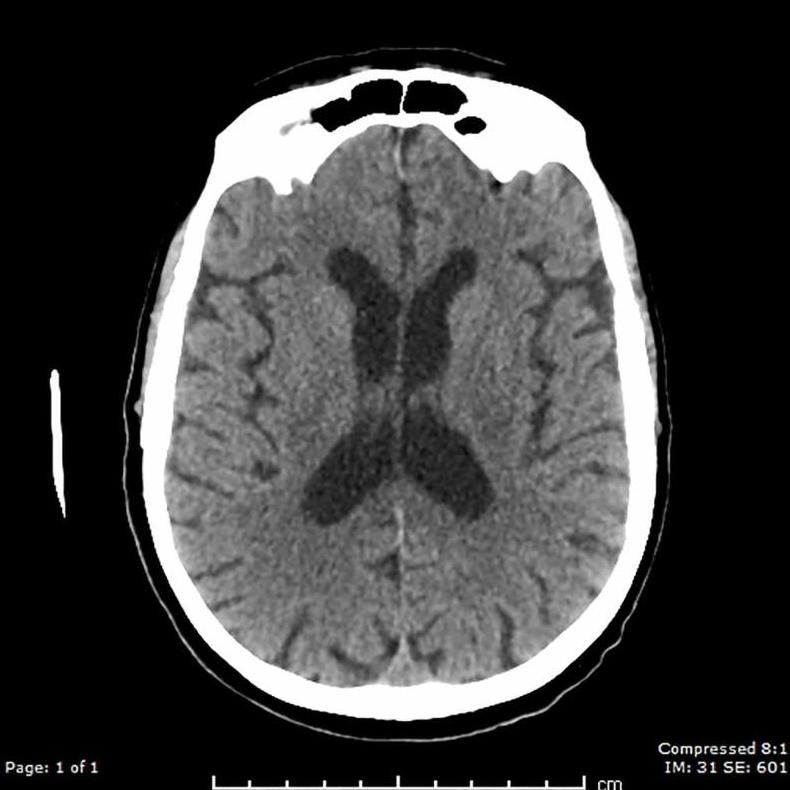}
& Findings / description only
& \textbf{Question:} Based on this image, which of the following best describes the primary imaging finding?

\textbf{Options:} A. Chronic diffuse cerebral atrophy (ex-vacuo ventriculomegaly);
B. Acute territorial ischemic infarction;
C. Acute intracranial hemorrhage;
D. Subdural hematoma with midline shift;
E. Normal noncontrast head CT. \\[0.8em]

\includegraphics[width=0.95\linewidth]{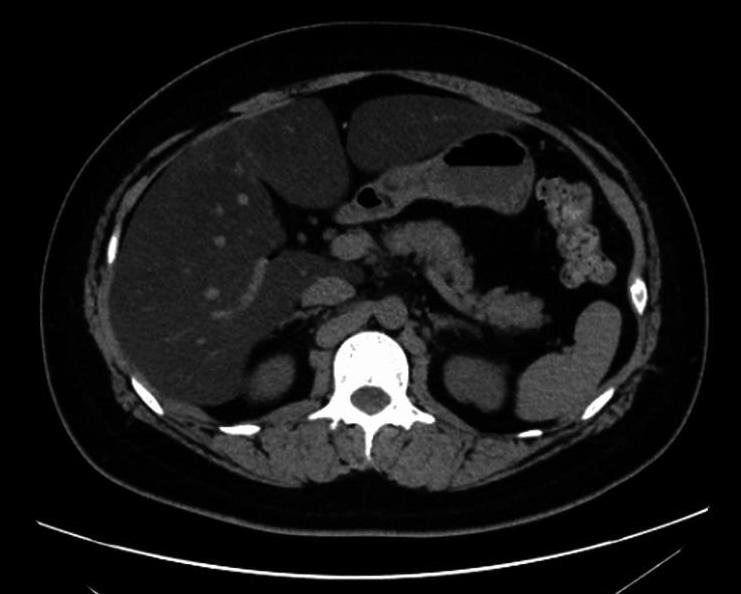}
& Anatomy / localization
& \textbf{Question:} Which organ occupies the majority of the left side of this axial abdominal CT image?

\textbf{Options:} A. Liver;
B. Spleen;
C. Left kidney;
D. Stomach. \\[0.8em]

\includegraphics[width=0.95\linewidth]{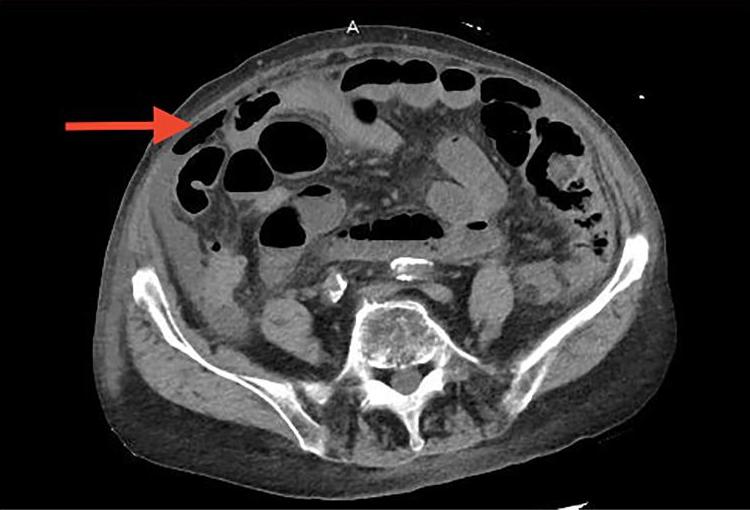}
& Annotation / marker
& \textbf{Question:} What does the red arrow indicate on this image?

\textbf{Options:} A. Simple renal cortical cyst;
B. Complex enhancing renal mass;
C. Enlarged para-aortic lymph node;
D. Normal renal parenchyma. \\[0.8em]

\includegraphics[width=0.95\linewidth]{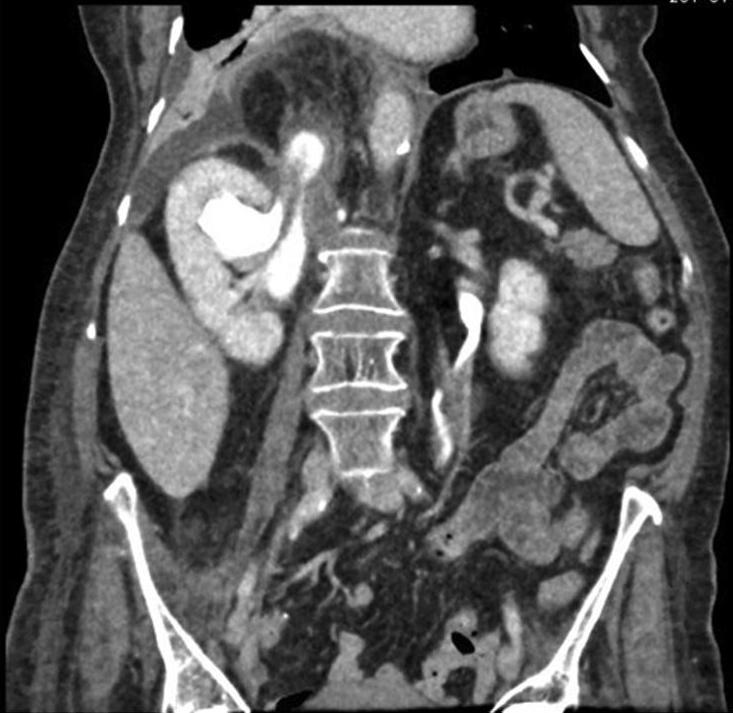}
& Spatial location
& \textbf{Question:} On the displayed image, in which region is the finding best localized?

\textbf{Options:} A. Upper left quadrant of the image;
B. Upper right quadrant of the image;
C. Lower left quadrant of the image;
D. Lower right quadrant of the image. \\[0.8em]

\includegraphics[width=0.95\linewidth]{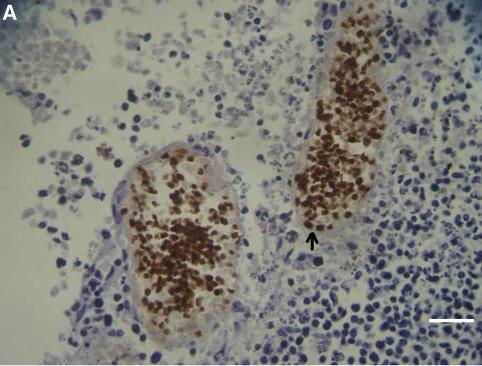}
& Counting
& \textbf{Question:} How many proliferating blood vessels are visible in this image?

\textbf{Options:} A. One;
B. Two;
C. Three;
D. Four or more. \\[0.8em]

\includegraphics[width=0.95\linewidth]{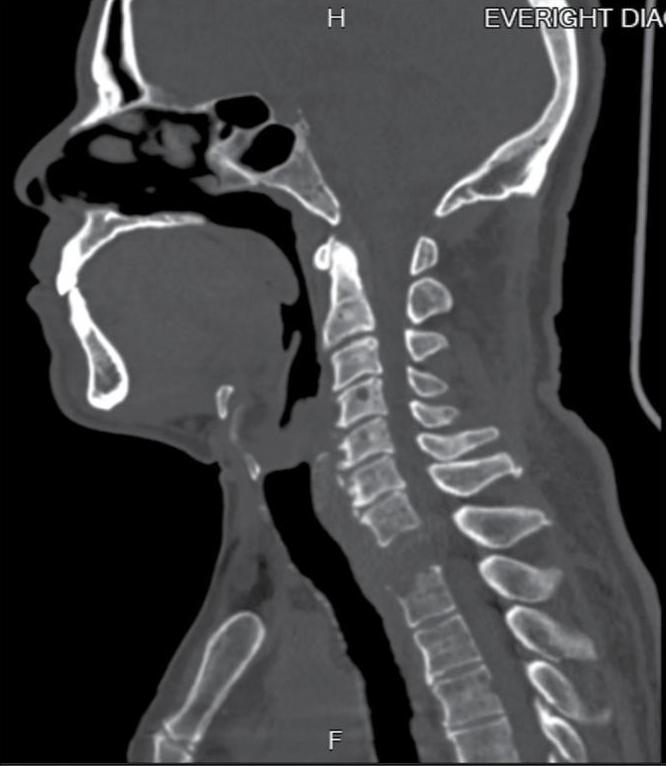}
& Normal vs.\ abnormal
& \textbf{Question:} Based on this image, are the findings normal or abnormal?

\textbf{Options:} A. Normal;
B. Abnormal. \\[0.8em]

\includegraphics[width=0.95\linewidth]{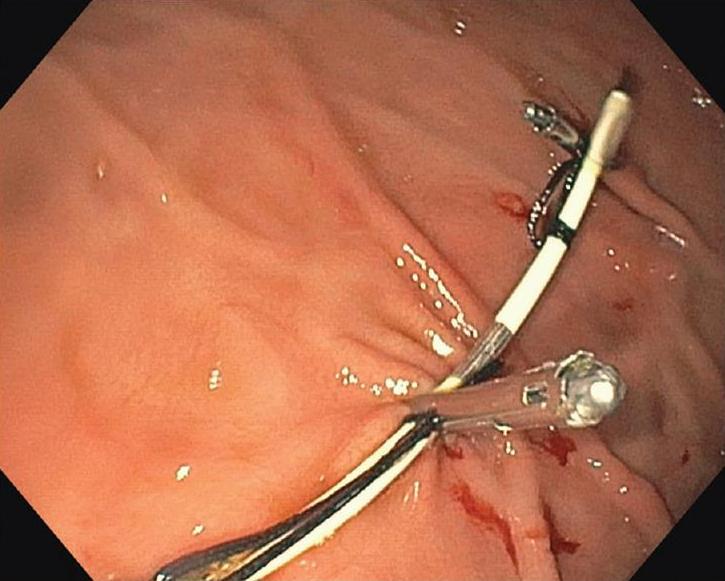}
& Diagnosis
& \textbf{Question:} Based on this image, which of the following is the most likely diagnosis?

\textbf{Options:} A. Temporary gastric electrical stimulation leads attached to mucosa;
B. Percutaneous endoscopic gastrostomy (PEG) tube;
C. Endoscopic hemostatic clips for bleeding;
D. Nasogastric tube in the stomach. \\[0.8em]

\includegraphics[width=0.95\linewidth]{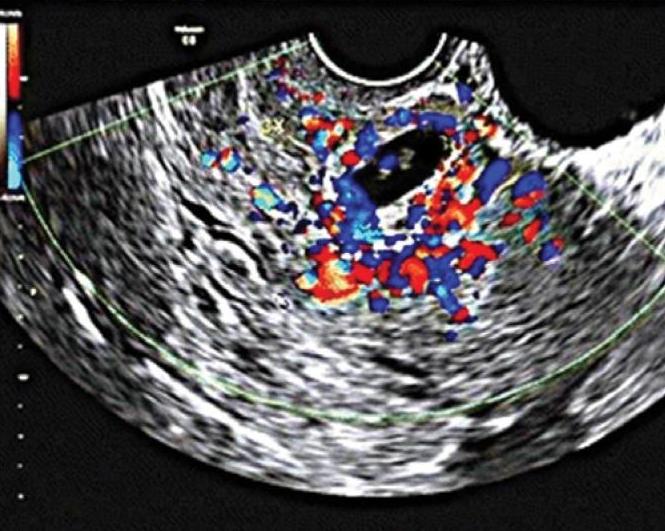}
& Mechanism / pathophysiology
& \textbf{Question:} Which pathophysiologic process best explains the vascular Doppler pattern seen on this image?

\textbf{Options:} A. Trophoblastic invasion of a cesarean scar with neovascularization;
B. Normal intrauterine implantation with low-resistance placentation;
C. Tubal ectopic pregnancy with surrounding hemorrhage;
D. Degenerating submucosal fibroid with peripheral hyperemia;
E. Retained products of conception causing inflammatory hyperemia. \\[0.8em]

\includegraphics[width=0.95\linewidth]{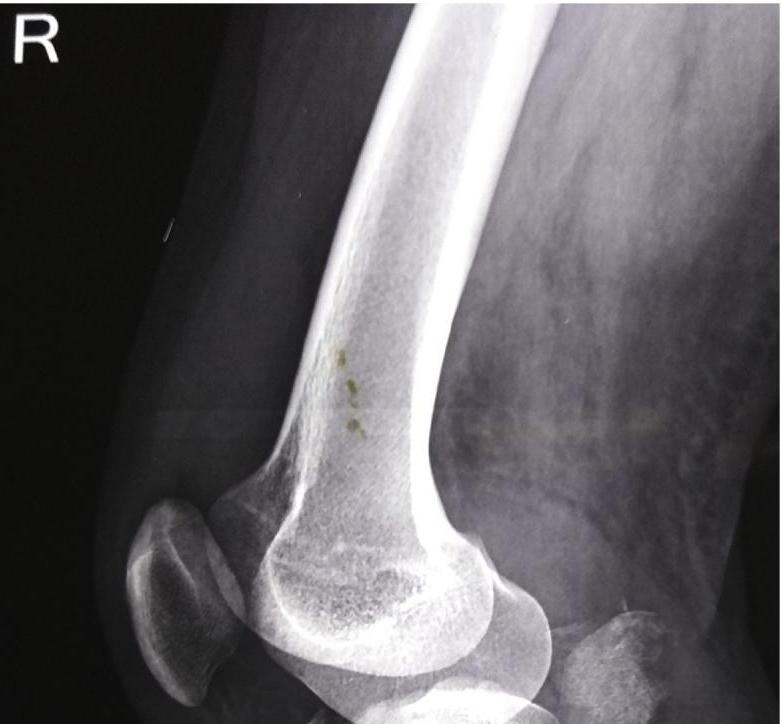}
& Differential diagnosis
& \textbf{Question:} Based on this image, is the abnormality more suggestive of an acute displaced lesion rather than chronic degenerative change?

\textbf{Options:} A. Yes;
B. No. \\[0.8em]

\includegraphics[width=0.95\linewidth]{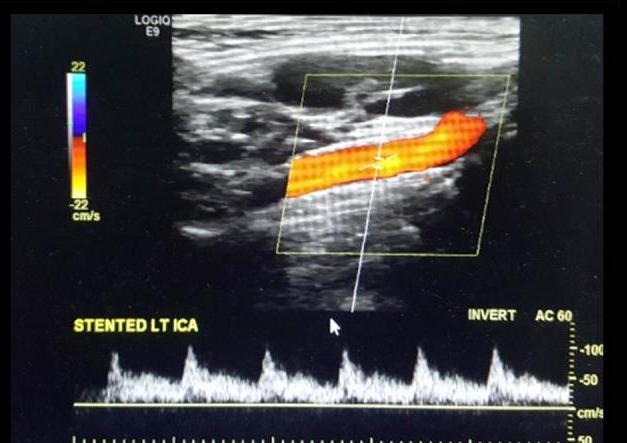}
& Severity grading
& \textbf{Question:} Based on this image, what is the severity grade?

\textbf{Options:} A. Mild ($<50\%$ stenosis);
B. Moderate ($50$--$69\%$ stenosis);
C. Severe ($\geq70\%$ stenosis);
D. Occluded. \\[0.8em]

\includegraphics[width=0.95\linewidth]{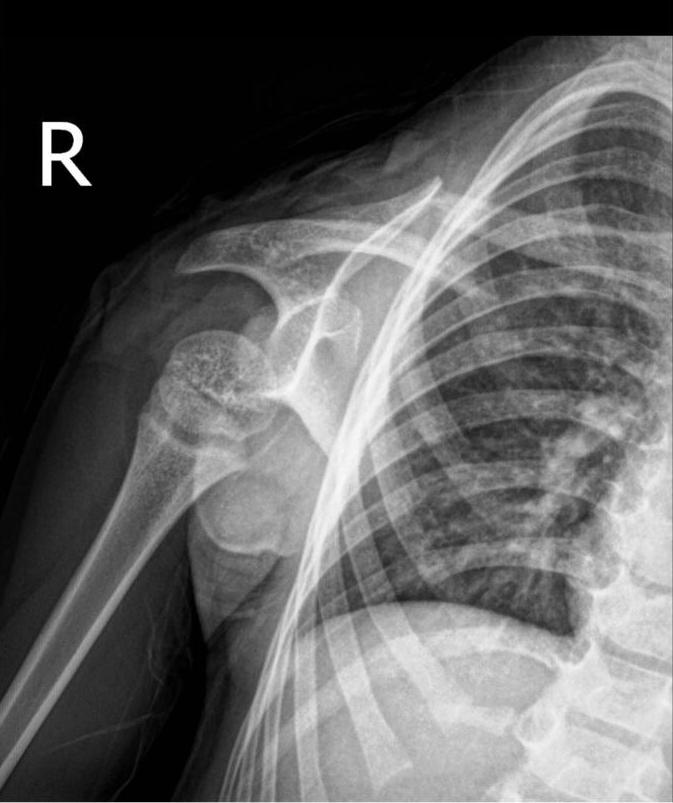}
& Next-step management
& \textbf{Question:} A patient presents after a fall with acute right shoulder pain and visible deformity. Neurovascular examination is intact and radiographs were obtained. Based on the imaging, what is the most appropriate next-step management?

\textbf{Options:} A. Urgent closed reduction of the shoulder;
B. Urgent open reduction and clavicle fixation;
C. Sling immobilization and outpatient follow-up;
D. Obtain advanced chest imaging before intervention. \\[0.8em]

\includegraphics[width=0.95\linewidth]{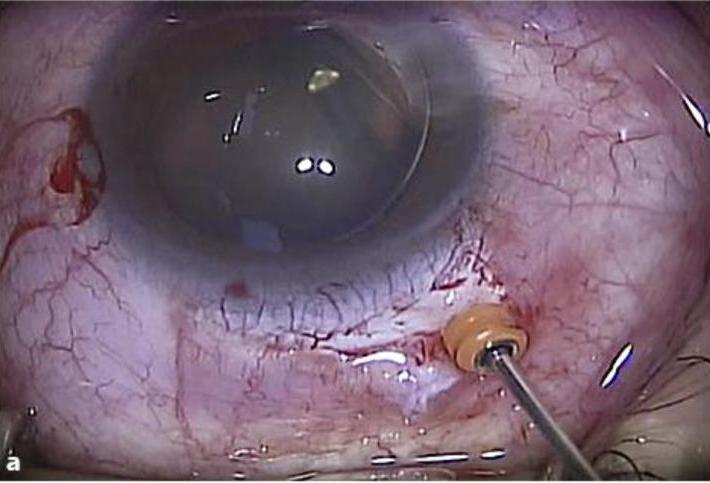}
& Surgical management
& \textbf{Question:} Mannitol and pharmacologic dilation were ineffective, so the team proceeded with surgical management that included partial pars plana dry vitrectomy, intracapsular lens extraction, and planned scleral fixation. Corneal paracenteses and viscoelastic injection were performed, and a sclerotomy with a 23-gauge microvitreoretinal blade 3 mm from the limbus was made with a trocar placed for instrumentation. Based on the image, which intraoperative step is being performed?

\textbf{Options:} A. Scleral trocar insertion with vitreous cutter;
B. Corneal paracentesis with viscoelastic injection;
C. Phacoemulsification of the lens;
D. Scleral fixation of intraocular lens;
E. Intracapsular lens extraction. \\[0.8em]

\includegraphics[width=0.95\linewidth]{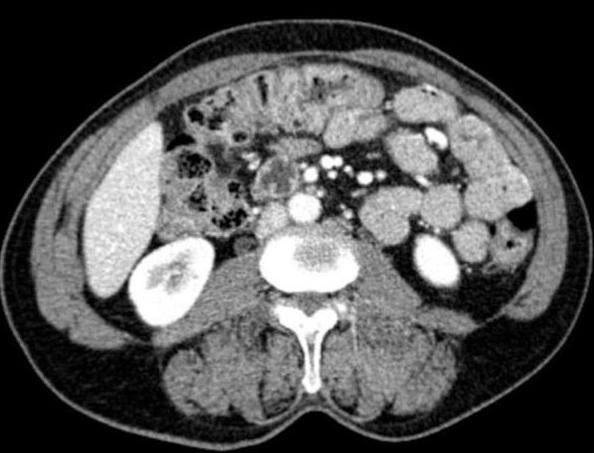}
& Drug therapy
& \textbf{Question:} A patient has received 10 months of oral amoxicillin for an abdominal inflammatory condition. She now undergoes repeat abdominal CT to assess treatment response. Based on this imaging, which of the following is the most appropriate pharmacologic action?

\textbf{Options:} A. Discontinue antibiotics;
B. Continue oral amoxicillin;
C. Switch to IV broad-spectrum antibiotics;
D. Add systemic corticosteroids;
E. Start empiric antifungal therapy. \\[0.8em]

\includegraphics[width=0.95\linewidth]{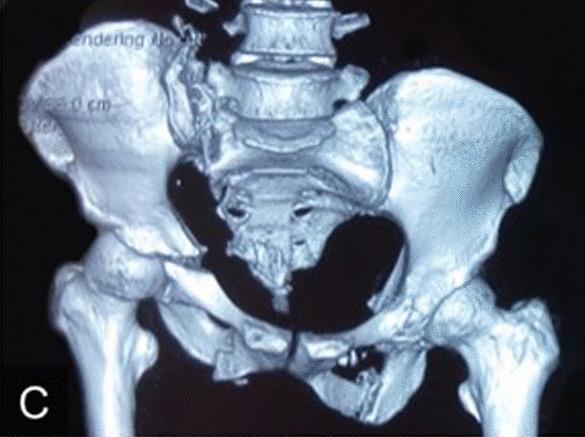}
& Complications
& \textbf{Question:} Several months after a pelvic fracture treated with attempted femoral skeletal traction that failed, a patient presents with progressive claudication. A three-dimensional CT reconstruction is provided. Which complication of the prior failed reduction best explains the clinical presentation and imaging?

\textbf{Options:} A. Sacral malunion with pelvic obliquity;
B. Pelvic ring nonunion with instability;
C. Deep pelvic osteomyelitis;
D. Avascular necrosis of the femoral head;
E. Heterotopic ossification causing nerve entrapment. \\[0.8em]

\includegraphics[width=0.95\linewidth]{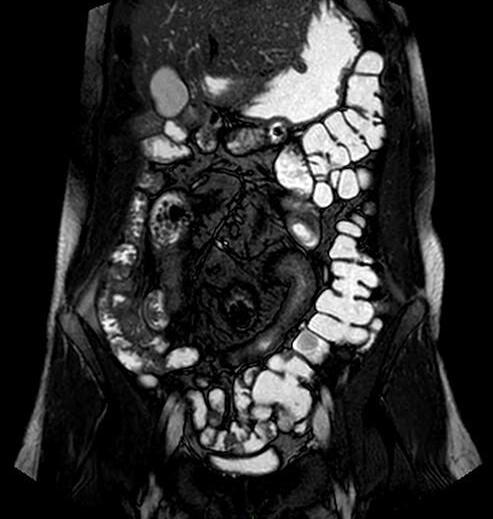}
& Symptoms / signs
& \textbf{Question:} A patient is admitted with worsening abdominal complaints. Prior MRI enterography from one year earlier is available for review. Based on the imaging, which symptom or clinical sign is most likely to be present?

\textbf{Options:} A. Bilious vomiting;
B. Progressive abdominal distension and obstipation;
C. Bright red rectal bleeding;
D. Painless jaundice. \\[0.8em]

\includegraphics[width=0.95\linewidth]{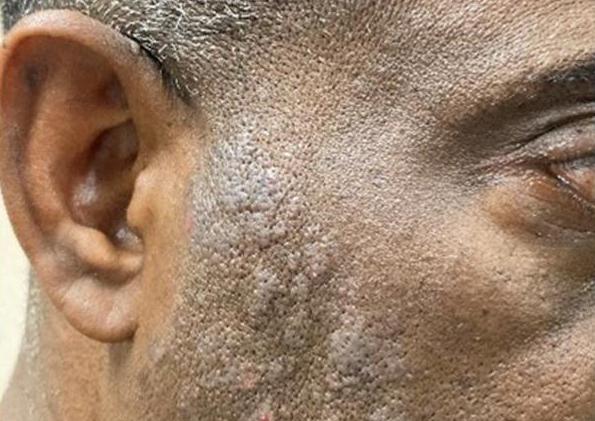}
& Safety / contraindications
& \textbf{Question:} A patient developed a skin eruption within 48 hours after using a commercial black hair dye; the product contained paraphenylenediamine. There are no systemic symptoms. Which of the following is the most appropriate safety recommendation?

\textbf{Options:} A. Administer intramuscular epinephrine;
B. Perform immediate desensitization exposure;
C. Avoid future hair dyes containing paraphenylenediamine;
D. Recommend using untested vegetable henna products;
E. Obtain patch testing before any future dye exposure. \\[0.8em]

\includegraphics[width=0.95\linewidth]{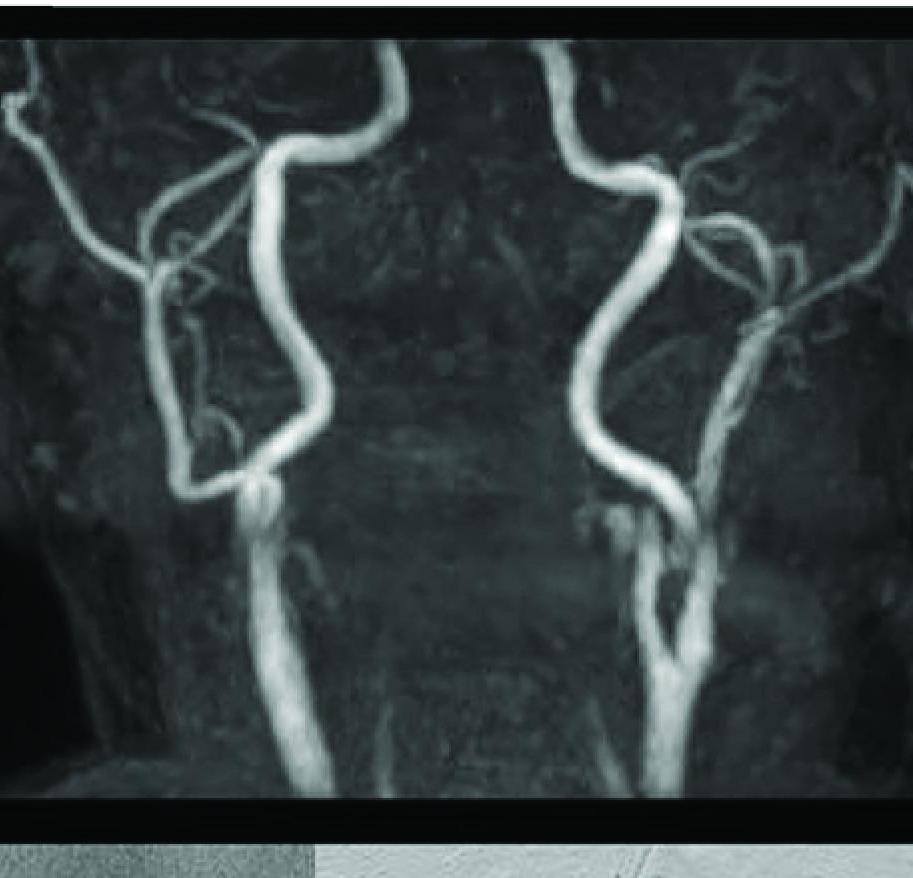}
& Next-step diagnostic test
& \textbf{Question:} An elderly patient underwent urgent endovascular cerebral arterial intervention with stent placement for acute neurologic deterioration. Post-procedure vascular imaging is available. Which additional diagnostic study is most appropriate to evaluate intracranial arterial patency and permit immediate endovascular therapy if required?

\textbf{Options:} A. Emergency diagnostic cerebral angiography (DSA);
B. CT angiography of head and neck;
C. Repeat non-contrast brain MRI;
D. Carotid duplex ultrasonography;
E. Transcranial Doppler ultrasound. \\[0.8em]

\end{longtable}

\endgroup

\begin{promptbox}[breakable]{Question Generation Prompt}
\label{prompt:question-generation}

\textbf{Role.}
You are a medical visual question answering dataset curator. Your task is to generate a high-quality, USMLE-style multiple-choice question from a medical image and its associated textual context.

\vspace{0.5em}
\noindent\textbf{Task.}
Given an image, its modality, sub-caption, textual context, target question category, and image scope, generate one clinically grounded multiple-choice question. The question must test the target category, require interpretation of the image, and be supported by the provided context. The answer format should be selected automatically based on the image, context, and target question category.

\vspace{0.5em}
\noindent\textbf{Input.}
The prompt is provided with the following fields:
\begin{itemize}
    \item \textbf{Modality:} the imaging or visual modality of the example.
    \item \textbf{Image:} the image used for the question.
    \item \textbf{Sub-caption:} the caption associated with the image.
    \item \textbf{Context:} the textual context associated with the image.
    \item \textbf{Target MCQ category:} the clinical task category to generate a question for.
    \item \textbf{Image scope:} whether the question refers to a subfigure or full figure.
    \item \textbf{Category-specific instructions:} additional category definitions and constraints used during prompting. These are included during generation but omitted here for space.
\end{itemize}

\vspace{0.5em}
\noindent\textbf{\_\_INVALID\_\_ gate.}
Before generating a question, first decide whether the example can support a valid MCQ for the target category. Return \verb|__INVALID__| instead of generating a question if any of the following conditions hold:

\begin{itemize}
    \item \textbf{Unsupported category:} the target category is not sufficiently supported by the provided image and context.
    
    \item \textbf{Insufficient rationale:} the context does not contain enough explanation or supporting evidence to justify why the correct answer is correct.
    
    \item \textbf{No single best answer:} the image or context is ambiguous, nondiagnostic, incomplete, or supports multiple defensible answers.
    
    \item \textbf{Missing required information:} the correct answer would require information not present in the provided image or context.
    
    \item \textbf{Image not necessary:} the question would be answerable from the textual context alone without interpreting the image.
    
    \item \textbf{Unseen visual dependency:} the question would require another panel, another timepoint, another imaging view, or any visual information not available in the provided input.
    
    \item \textbf{Stem leakage required:} a fair stem would need to reveal the diagnosis, finding, abnormality, location, appearance, or other case-specific visual information.
    
    \item \textbf{Invented information required:} the question would require adding patient history, imaging findings, laboratory values, diagnosis, treatment, prognosis, or other clinical details not present in the provided context.
    
    \item \textbf{Long-style failure:} a long-style question cannot be written using genuine non-visual clinical information from the context without describing the image or inventing details.
    
    \item \textbf{Category-specific failure:} a category-specific requirement is not satisfied; for example, an annotation question is requested but no visible non-letter marker is present.
\end{itemize}

\vspace{0.5em}
\noindent\textbf{General MCQ requirements.}
\begin{itemize}
    \item Generate exactly one MCQ for the target category.
    \item Test a single clear clinical or visual reasoning objective.
    \item Use a concise, clinically realistic, and professional stem.
    \item Ensure the correct answer is fully supported by the image and context.
    \item Ensure all distractors are plausible, homogeneous, and clearly inferior to the correct answer.
    \item Avoid trivia, overly rare edge cases, unnecessary details, unequal option lengths, overlapping choices, and unsupported absolute wording such as ``always'' or ``never''.
\end{itemize}

\vspace{0.5em}
\noindent\textbf{Image-use requirements.}
\begin{itemize}
    \item The question must require interpretation of the provided image.
    \item Use only the visual information available in the provided input.
    \item Do not depend on unseen panels, timepoints, views, source figures, or captions.
    \item Do not refer to source figure numbers, captions, sub-captions, panel letters, or the provided context in the question stem.
\end{itemize}

\vspace{0.5em}
\noindent\textbf{Stem privacy requirements.}
\begin{itemize}
    \item Do not reveal the diagnosis, disease, pathologic condition, or case-specific visual finding in the stem.
    \item Do not state or imply that a lesion, mass, opacity, abnormality, finding, structure, or diagnosis has already been identified, seen, demonstrated, detected, found, or shown.
    \item Do not describe visible image appearance in the stem, including location, laterality, size, morphology, pattern, grade, color, count, or other visual features.
    \item Generic phrases such as ``based on this image'' or ``based on the imaging'' are allowed only when they do not disclose what the image shows.
    \item Non-visual clinical information from the context may be used only if it does not make the answer obvious without interpreting the image.
\end{itemize}

\vspace{0.5em}
\noindent\textbf{Question style requirements.}
\begin{itemize}
    \item For \textbf{short} questions, use a minimal image-forward stem, preferably one sentence, without patient demographics, symptom narratives, or clinical vignette padding.
    \item For \textbf{long} questions, use only real non-visual clinical information from the supplied context.
    \item Never lengthen a stem by describing the image or adding unsupported clinical details.
\end{itemize}

\vspace{0.5em}
\noindent\textbf{Answer format selection.}
Choose the most appropriate answer format based on the target category, image content, and context:

\begin{itemize}
    \item Use \verb|standard| multiple choice with four or five options when the task requires selecting among several diagnoses, findings, anatomical structures, treatments, mechanisms, complications, or clinical decisions.
    \item Use \verb|binary_yesno| only when the most natural question asks whether a specific statement or feature is present or applicable. The choices must be exactly \verb|A = Yes| and \verb|B = No|.
    \item Use \verb|binary_truefalse| only when the most natural question asks whether a statement is true or false. The choices must be exactly \verb|A = True| and \verb|B = False|.
    \item Use \verb|binary_normal_abnormal| only for normal-versus-abnormal classification. The choices must be exactly \verb|A = Normal| and \verb|B = Abnormal|.
    \item Prefer \verb|standard| multiple choice when several plausible alternatives can be constructed fairly.
\end{itemize}

\vspace{0.5em}
\noindent\textbf{Category-specific requirements.}
\begin{itemize}
    \item For \textbf{Annotation / marker interpretation}, the image must contain a visible non-letter graphic marker such as an arrow, circle, box, star, bracket, outline, or pointer. The stem may refer to that marker, but must not use panel letters or reveal the diagnosis.
    
    \item For \textbf{Spatial location on image (quadrant / region)}, the stem may ask where a neutral referent, such as ``the finding'' or ``the abnormality,'' is located on the displayed image. The stem must not state the correct region or describe the finding.
    
    \item For all other categories, do not mention arrows, circles, boxes, labels, markers, or other annotations in the stem.
\end{itemize}

\vspace{0.5em}
\noindent\textbf{Output format.}
Return only a valid JSON object in the following format:

\begin{verbatim}
{
  "question": "...",
  "choices": {
    "A": "...",
    "B": "...",
    "C": "...",
    "D": "..."
  },
  "answer": "A",
  "image_scope": "subfigure"
}
\end{verbatim}

If the question is invalid, return exactly:

\begin{verbatim}
{
  "question": "__INVALID__",
  "choices": {
    "A": "N/A",
    "B": "N/A",
    "C": "N/A",
    "D": "N/A"
  },
  "answer": "N/A",
  "image_scope": "[REQUESTED_SCOPE]"
}
\end{verbatim}

Do not include markdown, comments, explanations, extra keys, or trailing commas.

\end{promptbox}


\subsection{Reasoning Traces Generation}
\label{app:reasoning}
The reasoning-generation stage (Section~\ref{sec:reasoning-generation}) produces a source-grounded rationale for each accepted question--answer pair using a two-stage draft-then-refine procedure.

\textbf{Stage~1: medical-VLM draft.} In this step Octomed~\cite{ossowski2025octomed} which is a medical foundation reasonig model was used. This model receives the image, source context, question, and correct answer and produces an initial draft. The draft is conditioned on the source context to keep the explanation grounded in the medical paper rather than in parametric medical knowledge. The medical VLM is used as an evidence-organization scaffold by identifying the visual target, linking it to the question, and assembling supporting source statements. Minor background links may be introduced only when needed for coherence and only if they do not add unsupported case-specific claims. The full draft prompt is shown in Prompt~\ref{prompt:reasoning_draft}.

\begin{promptbox}[breakable]{Reasoning Draft Generation Prompt}
\label{prompt:reasoning_draft}

\textbf{Role.}
You are a medical visual reasoning assistant. Given a medical image, context, question, and correct answer, write a reasoning trace that explains why the answer is correct.

\vspace{0.4em}
\noindent\textbf{Task.}
Generate the reasoning inside \verb|<think>|. The reasoning must connect the image evidence and relevant clinical interpretation to the provided answer.

\vspace{0.4em}
\noindent\textbf{Requirements.}
\begin{itemize}
    \item Briefly state what must be checked in the image and context.
    \item Include a \textbf{Perception} part describing the key visual findings needed to answer.
    \item Include a \textbf{Clinical interpretation} part explaining what those findings mean clinically and discuss the given medical context in the question.
    \item Focus only on clinically relevant findings; mention technical details only if needed.
    \item End with a summary linking the image findings and clinical interpretation to the answer.
\end{itemize}

\vspace{0.4em}
\noindent\textbf{Output format.}
Return only:

\begin{Verbatim}[breaklines, fontsize=\small]
<think>
[Brief plan.]

Perception:
[Key image findings.]

Clinical interpretation:
[What the findings mean clinically and why they support the answer.]

[Brief final summary.]
</think>

<answer>[Final answer option letter]</answer>
\end{Verbatim}

\end{promptbox}

\textbf{Stage~2: Reasonign refinement}
The initial reasoning draft is then refined using \texttt{gpt-5-mini}. The refiner verifies the draft against the image, question, correct answer, caption, and in-text references. Claims that are unsupported by the image or source context are removed, as are article-level digressions and details that are not needed to justify the answer. When the draft contains useful but poorly organized evidence, the refiner restructures it into a concise rationale; when it contains details that are inconsistent with or absent from the source, those details are corrected or discarded. The full refinement prompt is shown in Prompt~\ref{prompt:reasoning_refine}.

\textbf{Trace format.}
Each retained trace is rewritten into a standardized evidence-to-answer format. The format separates the reasoning into three connected components:
\begin{enumerate}
    \item \textbf{Perception.} The trace first states the image-grounded observations needed to answer the question. This includes only clinically relevant visual evidence, such as the modality, anatomical region, visible abnormality, spatial pattern, morphology, signal, density, uptake, or microscopic appearance, depending on the image type.
    
    \item \textbf{Clinical interpretation and medical knowledge.} The trace then explains how the perceptual evidence should be interpreted clinically or biomedically. This step uses relevant medical knowledge, together with the source context, to connect the observed findings to the diagnosis, mechanism, management decision, anatomical interpretation, risk assessment, or other task-specific answer.
    
    \item \textbf{Answer justification.} The trace ends with a concise justification that explicitly links the key perceptual evidence, clinical interpretation, and relevant medical knowledge to the selected answer.
\end{enumerate}
This structure keeps the rationale focused on the path from image evidence, through clinically grounded medical knowledge, to the final answer, while avoiding unsupported speculation, unnecessary background information, or reasoning based primarily on eliminating answer choices.

\begin{promptbox}[breakable]{Reasoning Refinement Prompt}
\label{prompt:reasoning_refine}

\textbf{Role.}
You are a medical visual reasoning editor. Your task is to refine an initial reasoning draft for a medical multiple-choice question into a grounded, coherent reasoning trace.

\vspace{0.4em}
\noindent\textbf{Task.}
Given the image, image context, question, answer options, target answer, and draft reasoning, rewrite the draft so that it clearly explains why the target answer is correct. The refined trace must connect image-based perception, patient clinical context, clinical interpretation, and relevant medical knowledge to the final answer.

\vspace{0.4em}
\noindent\textbf{Inputs.}
The prompt is provided with:
\begin{itemize}
    \item \textbf{Image:} the medical image used to answer the question.
    \item \textbf{Image context:} the source context associated with the image, used only to verify and ground the reasoning.
    \item \textbf{Question and options:} the multiple-choice item to be answered.
    \item \textbf{Target answer:} the answer letter that the refined reasoning must support.
    \item \textbf{Draft reasoning:} the initial reasoning trace to be improved.
\end{itemize}

\vspace{0.4em}
\noindent\textbf{Grounding rules.}
\begin{itemize}
    \item Use only information supported by the image, image context, question, options, and target answer.
    \item Use the image context to verify that clinical claims, interpretation, and medical knowledge in the reasoning are consistent with the source.
    \item Preserve useful image-grounded observations from the draft when they are supported by the image or image context.
    \item Remove unsupported claims, invented clinical details, unnecessary background, and article-level digressions.
    \item Do not explicitly mention or quote the image context, caption, report, source text, article, or prior description in the final reasoning.
    \item Do not mention arrows, boxes, labels, markers, panel letters, or subfigure identifiers unless the question explicitly requires them.
    \item Do not state that a target answer was provided.
\end{itemize}

\vspace{0.4em}
\noindent\textbf{Trace structure.}
Inside \verb|<think>|, write a concise but complete reasoning trace with the following components:

\begin{itemize}
    \item Begin with a brief statement of what must be checked in the image and question.
    \item Include a labeled \textbf{Perception} part describing the key image findings needed to answer.
    \item Include a labeled \textbf{Clinical context} part summarizing the patient information provided in the question that is relevant to the answer.
    \item Include a labeled \textbf{Clinical interpretation and medical knowledge} part explaining how the image findings and patient context should be interpreted clinically, and how relevant medical knowledge supports the answer.
    \item End with a short answer justification that links the perceptual evidence, clinical context, and medical interpretation to the selected answer.
\end{itemize}

\vspace{0.4em}
\noindent\textbf{Style rules.}
\begin{itemize}
    \item Write in a clear, step-by-step clinical reasoning tone.
    \item Be specific enough to explain the answer, but avoid unnecessary repetition or filler.
    \item Do not compress the reasoning into a single short verdict.
    \item Do not reason mainly by eliminating answer choices; compare options only when needed for clarity.
    \item Avoid hedging unless genuine uncertainty is unavoidable.
    \item Do not add unsupported details to make the reasoning sound more expert.
\end{itemize}

\vspace{0.4em}
\noindent\textbf{Output format.}
Return only:

\begin{verbatim}
<think>
[Final reasoning trace explaining how to use
the image, questions and options to get to the final answer.]
</think>
<answer>X</answer>
\end{verbatim}

where \verb|X| is the target answer letter.

\end{promptbox}

\subsubsection{Final reasoning verification}
\label{app:reasoning-verification}

 

\begin{figure}[t]
\centering

\begin{tcolorbox}[
    colback=gray!3,
    colframe=gray!45,
    boxrule=0.4pt,
    arc=2mm,
    left=2mm,
    right=2mm,
    top=2mm,
    bottom=2mm
]
\centering
\includegraphics[width=0.70\linewidth]{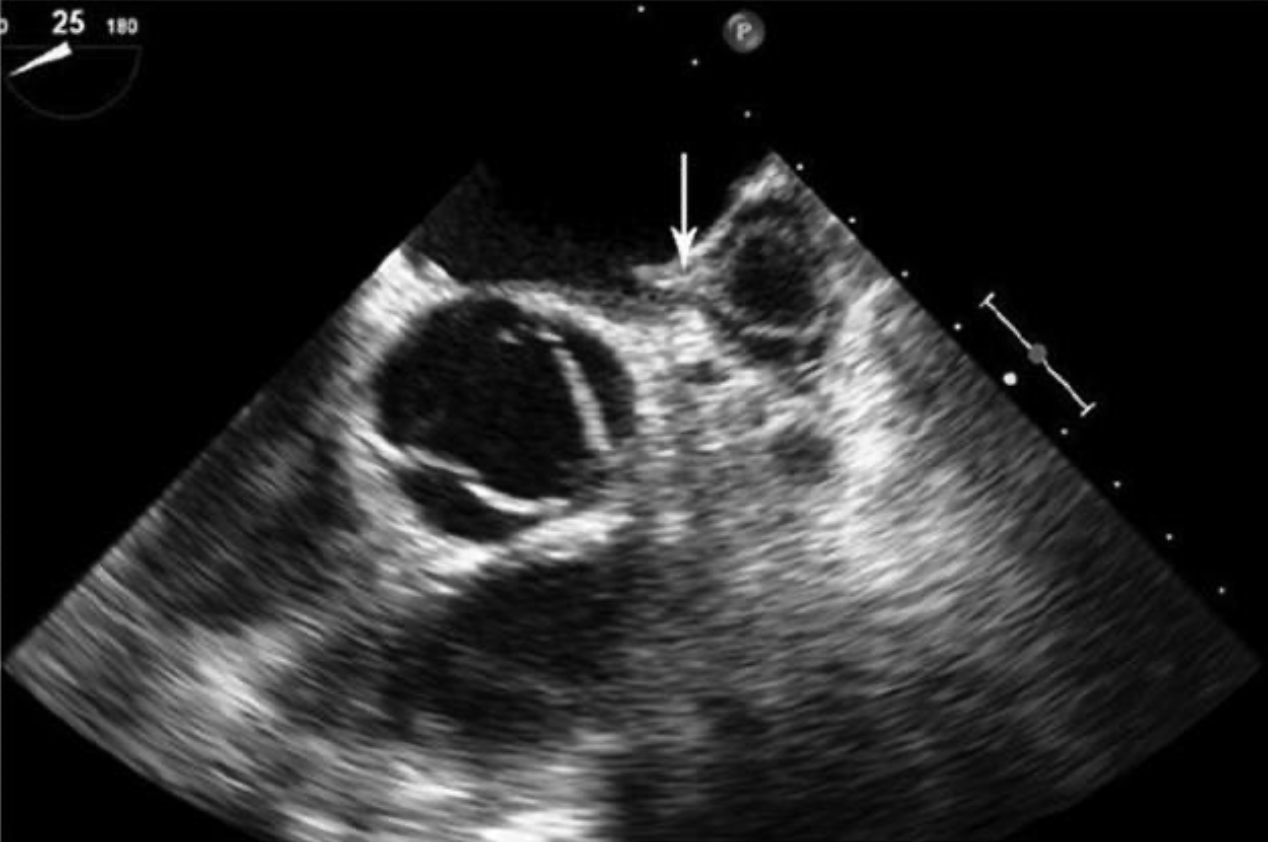}

\vspace{0.7em}

\begin{minipage}{0.94\linewidth}
\small

\textbf{Question.}
One month after initial presentation, the patient experienced two brief episodes of transient left-eye vision loss. Atrial fibrillation had recurred and transesophageal echocardiography was obtained, as shown above. Based on the imaging, which statement best describes the patient's short-term prognosis?

\vspace{0.4em}

\begin{tabular}{@{}ll@{}}
A. & High immediate risk of systemic embolization \\
B. & Low short-term embolic risk with outpatient therapy \\
C. & Predominant risk of progressive heart failure \\
D. & No change in short-term prognosis from baseline \\
\end{tabular}

\vspace{1.8em}

\textbf{Reasoning} First, I examine the transesophageal echocardiographic image. I identify the left atrial appendage (an anechoic cavity) and an echogenic, rounded mass at the appendage ostium pointed out by the arrow. The mass projects from the appendage toward the left atrium, with a narrow attachment consistent with a pedunculated intracavitary lesion.
\\

Next, I relate these imaging findings to the clinical question: the patient has recurrent atrial fibrillation and recent transient left-eye vision loss. An intracavitary mass in the left atrial appendage that projects into the left atrium in the setting of AF is anatomically well positioned to embolize to the systemic (cerebral/ocular) circulation.
\\

Then I consider how the answer choices align with those observations. A mass in the left atrial appendage that protrudes toward the left atrium represents an immediate source for systemic emboli, so the short-term embolic risk is increased. This contrasts with options that imply low short-term embolic risk, predominant heart-failure risk, or no change in short-term prognosis, none of which match the imaging appearance of a potential embolic source.
\\

In summary, the image shows an LAA intracavitary mass projecting into the left atrium in a patient with AF and transient visual ischemic events, which supports a high immediate risk of systemic embolization.
\\

\vspace{0.4em}
\textbf{Answer:} A. High immediate risk of systemic embolization.

\end{minipage}
\end{tcolorbox}

\caption{
Qualitative example of image-grounded medical reasoning. The transesophageal echocardiographic image shows an echogenic intracavitary mass near the left atrial appendage region, which, in the setting of recurrent atrial fibrillation and transient visual ischemic symptoms, supports high immediate risk of systemic embolization.
}
\label{fig:tee_laa_reasoning_example}
\end{figure}

\begin{figure}[t]
\centering

\begin{tcolorbox}[
colback=gray!3,
colframe=gray!45,
boxrule=0.4pt,
arc=2mm,
left=2mm,
right=2mm,
top=2mm,
bottom=2mm
]
\centering
\includegraphics[width=0.6\linewidth]{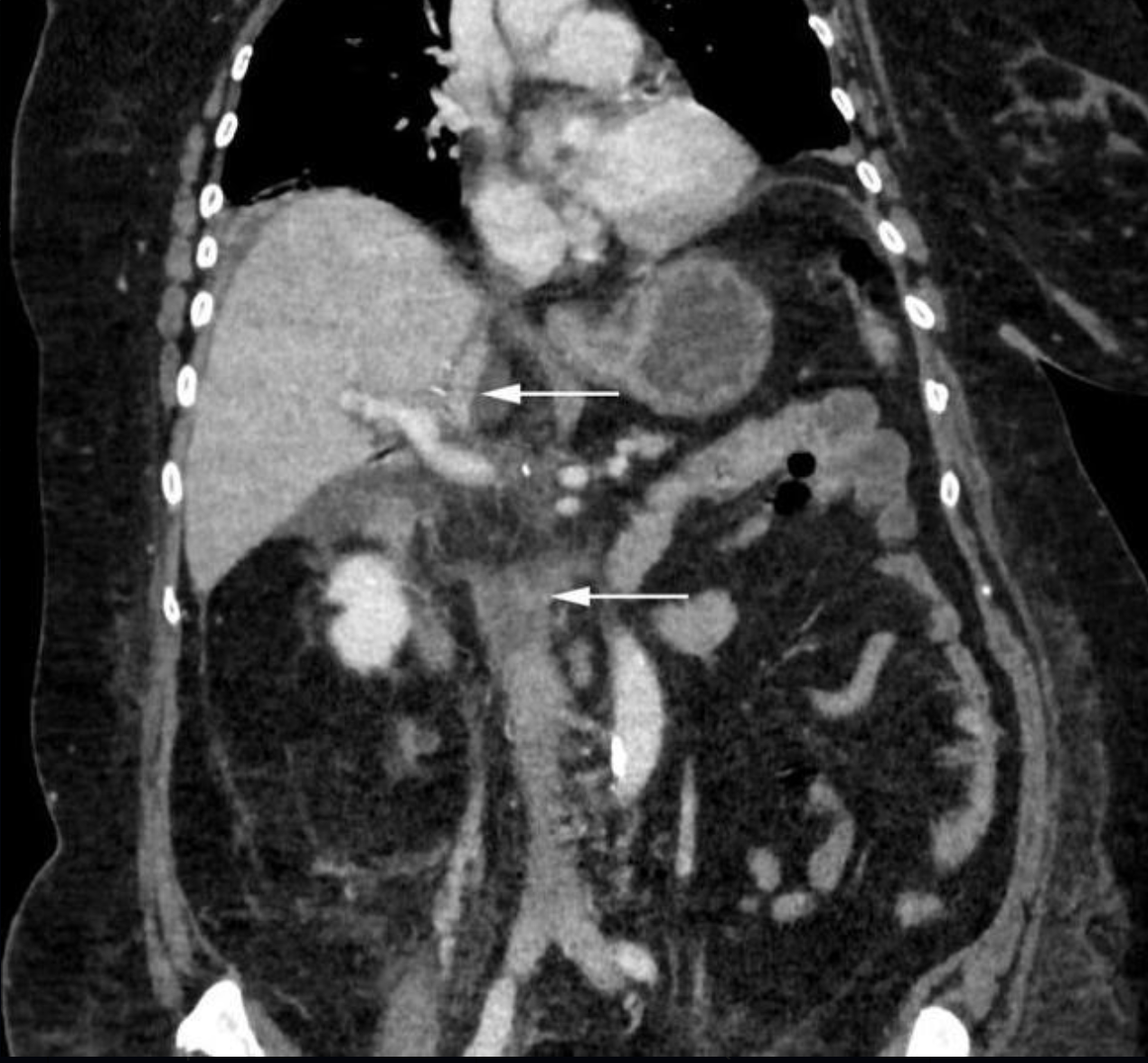}

\vspace{0.7em}

\begin{minipage}{0.94\linewidth}
\small

\textbf{Question.}
A patient underwent retroperitoneal tumor resection and a postoperative abdominal CT scan was obtained to evaluate for complications. Based on the imaging, which postoperative complication is most likely?

\vspace{0.4em}

\begin{enumerate}[label=\Alph*., leftmargin=1.6em, itemsep=0.15em, topsep=0.2em]
\item Iatrogenic inferior vena cava ligation/transection
\item Postoperative intra-abdominal abscess
\item Portal vein thrombosis
\item Ureteral transection
\item Arterial pseudoaneurysm
\end{enumerate}

\vspace{1.8em}

\textbf{Reasoning} First, I examine the provided coronal contrast-enhanced abdominal CT image for postoperative changes and abnormalities. I note a central retroperitoneal vascular structure along the expected course of the inferior vena cava that appears interrupted: there are two contrast-opacified blind-ending segments, or stumps, separated by a gap where continuity of the vessel is lost, with arrows pointing to these ends. There is no obvious peritoneal fluid collection with gas or a rim-enhancing fluid cavity to suggest an intra-abdominal abscess, and I do not see a focal contrast-filled arterial outpouching that would indicate a pseudoaneurysm.
\\

Next, I relate these imaging findings to the question, which asks which postoperative complication is most likely. The focal discontinuity of the main retroperitoneal venous conduit with proximal and distal stumps is most consistent with surgical ligation or transection of that vein. This appearance specifically involves the inferior vena cava along its expected course.
\\

Then, I consider how the other options line up with the image. Portal vein thrombosis would affect the portal venous branches near the liver rather than show a gap in the retroperitoneal caval column. Ureteral transection would be suggested by urinary leakage or hydronephrosis rather than an interrupted large-caliber central vein. An arterial pseudoaneurysm would appear as a focal enhancing outpouching, which is not present here.
\\

In summary, the image shows focal discontinuity of the inferior vena cava with blind-ending contrast-opacified stumps after retroperitoneal tumor resection, which supports iatrogenic inferior vena cava ligation or transection.
\\

\vspace{0.4em}
\textbf{Answer:} A. Iatrogenic inferior vena cava ligation/transection.

\end{minipage}
\end{tcolorbox}

\caption{
Qualitative example of image-grounded medical reasoning. The postoperative coronal contrast-enhanced abdominal CT image shows focal discontinuity of the inferior vena cava with blind-ending contrast-opacified segments, supporting iatrogenic inferior vena cava ligation or transection rather than abscess, portal vein thrombosis, ureteral injury, or arterial pseudoaneurysm.
}
\label{fig:ivc_transection_reasoning_example}
\end{figure}

\begin{figure}[t]
\centering

\begin{tcolorbox}[
    colback=gray!3,
    colframe=gray!45,
    boxrule=0.4pt,
    arc=2mm,
    left=2mm,
    right=2mm,
    top=2mm,
    bottom=2mm
]
\centering
\includegraphics[width=0.78\linewidth]{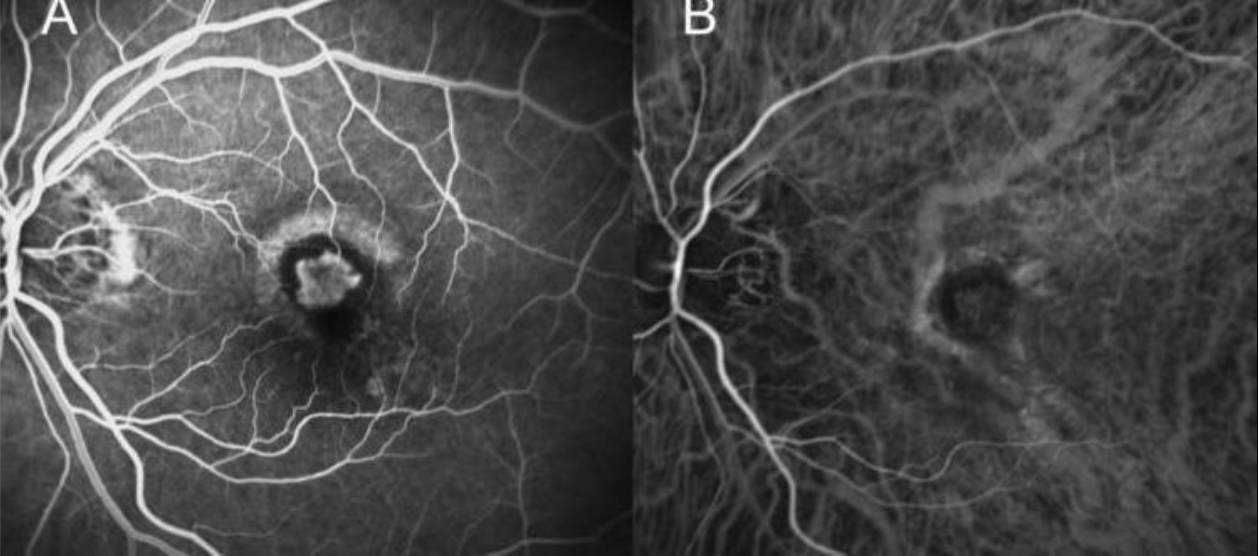}

\vspace{0.7em}

\begin{minipage}{0.94\linewidth}
\small

\textbf{Question.}
A patient with new central visual complaints underwent fluorescein and indocyanine green angiography; representative images are shown. Based on the imaging, which initial management is most appropriate?

\vspace{0.4em}

\begin{enumerate}[label=\Alph*., leftmargin=1.6em, itemsep=0.15em, topsep=0.2em]
    \item Intravitreal anti-VEGF injection
    \item Photodynamic therapy with verteporfin
    \item Focal thermal laser photocoagulation
    \item Systemic corticosteroid therapy
    \item Observation with interval follow-up
\end{enumerate}

\vspace{1.8em}

\textbf{Reasoning} First, I examine the fluorescein angiogram (panel A). I see a well-demarcated, nodular area of early central hyperfluorescence located at the macula, with a darker rim surrounding it. This appearance indicates an angiographically visible vascular membrane centered on the fovea.
\\

Next, I inspect the indocyanine green image (panel B). There is a corresponding abnormal vascular network in the macular region, seen as a focal plaque of altered ICG signal, that spatially matches the fluorescein lesion. This supports the presence of an active macular neovascular membrane rather than a nonvascular lesion.
\\

Then I relate these imaging findings to the clinical question. The patient has new central visual complaints, and the imaging shows an active, vascularized lesion directly involving the macula. This combination implies risk to central vision if left untreated.
\\

Now I consider the management options in light of those observations. Intravitreal anti-VEGF therapy targets intraocular neovascular activity and is an appropriate initial intervention for an active macular neovascular membrane. Focal thermal laser would risk destroying central retinal tissue when the lesion is subfoveal or juxtafoveal, and is therefore unsuitable for a central vascular membrane. Photodynamic therapy can be used selectively for some choroidal neovascular lesions but is not the immediate first-line choice for a symptomatic, angiographically active macular neovascular membrane seen here. Systemic corticosteroids do not address a focal vascular membrane, and simple observation would leave an active central lesion untreated despite new symptoms.
\\

In summary, the fluorescein and indocyanine green angiograms show an active macular neovascular membrane corresponding to the patient's new central visual symptoms, which supports intravitreal anti-VEGF injection as the most appropriate initial management.
\\

\vspace{0.4em}
\textbf{Answer:} A. Intravitreal anti-VEGF injection.

\end{minipage}
\end{tcolorbox}

\caption{
Qualitative example of image-grounded medical reasoning. The fluorescein and indocyanine green angiograms show an active vascular lesion in the macular region consistent with a macular neovascular membrane, which in a patient with new central visual complaints supports initial treatment with intravitreal anti-VEGF therapy.
}
\label{fig:macular_neovascular_reasoning_example}
\end{figure}

\textbf{Stage~3: verification with \texttt{gpt-5}.}
As a final quality-control step, we use \texttt{gpt-5} as a verifier to determine whether each generated question--answer pair and its reasoning trace should be retained. The verifier is given the image, source context, question, answer options, selected answer, and reasoning trace. Its role is not to rewrite the example, but to judge whether the question and trace are grounded in the available evidence and useful for explaining the answer. The full verification prompt is shown in Prompt~\ref{prompt:reasoning_verify}.

\textbf{Decision rules.}
A question--reasoning pair is rejected if it fails any of the following three criteria:

\begin{itemize}
    \item \textbf{Source consistency.}
    The question, selected answer, and reasoning trace must be consistent with the image and source context. Any claim that is absent from, stronger than, or contradictory to the available evidence is grounds for rejection. Common failures include invented anatomical landmarks, fabricated measurements, incorrect modality or laterality, unsupported lesion descriptions, clinical claims that contradict the caption or source context, and reasoning that imports article-level information not tied to the displayed image.

    \item \textbf{Answer justification.}
    The reasoning trace must explain how the selected answer follows from the visual and contextual evidence. It is not sufficient to restate the question, mention isolated correct facts, or simply state the answer. Common failures include traces that identify relevant concepts but do not connect them to the chosen option, traces that provide a conclusion without inference, and rationales that could support multiple answer choices equally well.

    \item \textbf{Reasoning utility.}
    The reasoning trace must contain information that is useful for reaching the answer. It should focus on the image evidence, relevant patient context, and clinical or biomedical interpretation needed for the specific question. Common failures include generic medical background, article-level study results unrelated to the displayed image, excessive option-by-option elimination, unsupported speculation, and verbose explanations that obscure the key evidence.
\end{itemize}

Only examples that satisfy all three criteria are retained. This conservative filtering step prioritizes grounded, answer-directed reasoning over coverage.

\begin{promptbox}[breakable]{Reasoning Verification Prompt}
\label{prompt:reasoning_verify}

\textbf{Role.}
You are a medical visual reasoning verifier. Your task is to judge whether a generated medical VQA example is grounded in the provided evidence and useful for explaining the selected answer.

\vspace{0.4em}
\noindent\textbf{Task.}
Given the image, source context, question, answer options, selected answer, and reasoning trace, decide whether the example should be retained or rejected. Do not rewrite the question or reasoning. Only verify whether the question, selected answer, and reasoning trace are supported by the available evidence.

\vspace{0.4em}
\noindent\textbf{Inputs.}
The prompt is provided with:
\begin{itemize}
    \item \textbf{Image:} the medical image used in the question.
    \item \textbf{Source context:} the caption, in-text reference, or surrounding text associated with the image.
    \item \textbf{Question and options:} the generated multiple-choice item.
    \item \textbf{Selected answer:} the answer chosen for the item.
    \item \textbf{Reasoning trace:} the generated rationale explaining the selected answer.
\end{itemize}

\vspace{0.4em}
\noindent\textbf{Verification criteria.}
Reject the example if it fails any of the following three criteria:

\begin{itemize}
    \item \textbf{Source consistency.}
    The question, selected answer, and reasoning trace must be consistent with the image and source context. Reject if any claim is unsupported, exaggerated beyond the evidence, or contradictory to the source. This includes invented anatomy, fabricated measurements, incorrect modality or laterality, unsupported visual findings, invented clinical details, or article-level information not tied to the displayed image.

    \item \textbf{Answer justification.}
    The reasoning trace must explain how the selected answer follows from the visual and contextual evidence. Reject if the trace merely restates the question, states the answer without explanation, lists isolated facts without connecting them to the answer, relies mainly on option elimination, or gives a rationale that could support more than one option.

    \item \textbf{Reasoning utility.}
    The reasoning trace must contain information that is useful for reaching the selected answer. Reject if the trace is mostly generic medical background, irrelevant source details, unsupported speculation, unnecessary article-level discussion, overly verbose rule-outs, or content that does not help justify the answer for this specific image--question pair.
\end{itemize}

\vspace{0.4em}
\noindent\textbf{Decision rule.}
Accept the example only if the question, answer, and reasoning trace satisfy all three criteria. If the example is partially grounded, ambiguous, medically questionable, or not clearly useful for explaining the selected answer, reject it.

\vspace{0.4em}
\noindent\textbf{Output format.}
Return only a valid JSON object:

\begin{verbatim}
{
  "decision": "accept" or "reject",
  "failed_criteria": [
    "source_consistency",
    "answer_justification",
    "reasoning_utility"
  ],
  "reason": "Brief explanation of the decision."
}
\end{verbatim}

If the example is accepted, return an empty list for \verb|failed_criteria|. If rejected, include only the failed criteria. Do not include markdown, comments, explanations outside the JSON, or extra keys.

\end{promptbox}


\subsubsection{\dataset~Bench: reasoning-trace evaluation}
\label{app:reasoning_trace_eval}
 
\paragraph{Reasoning-trace checklist construction.}
For each example, we convert the reference reasoning trace into a compact checklist of
axis-specific unit claims. The checklist is generated with \texttt{gpt-5-mini} from the
question stem, answer options, reference answer, and reference reasoning trace. The model
is instructed to extract a non-redundant set of atomic units that together cover the
distinct forward steps used by the reference trace to justify the answer.

Each unit corresponds to one required component of the reasoning trace and is assigned to
exactly one of three axes: \emph{perception}, \emph{medical knowledge}, or
\emph{rationale}.

\begin{itemize}
    \item \textbf{Perception units} capture image-grounded evidence used by the trace,
    such as modality, anatomical region, lesion or abnormality appearance, markers,
    counts, measurements, spatial patterns, and stated absences. Modifiers that describe
    the same finding, such as size, density, margin, or laterality, are grouped into the
    same unit.

    \item \textbf{Medical-knowledge units} capture general clinical facts used by the
    explanation. These are facts that remain true independently of the present case.

    \item \textbf{Rationale units} capture case-specific inferential links that connect
    the visual evidence and clinical knowledge to the answer. These include diagnostic
    conclusions, differential exclusion, causal or mechanistic links, next-step reasoning,
    treatment reasoning, prognosis, and severity judgments.
\end{itemize}

Units that merely restate the question stem or answer options are excluded. When the
reference trace argues against a specific option, the argument is rewritten as a positive
forward claim whenever possible, for example as a rule-out inference for a competing
diagnosis rather than as an option-letter comparison.

\paragraph{Presence and correctness probes.}
Each unit in the checklist is evaluated with two complementary probes: a \emph{presence} probe and a \emph{correctness} probe. The presence probe asks
whether the candidate trace engages with the topic of the unit at all. For example, for a
unit about a cystic pancreatic tail lesion, the presence probe asks whether the response
discusses that visual finding, regardless of whether it describes the finding correctly.
Thus, a trace can receive presence credit for raising the relevant concept even if its
conclusion disagrees with the reference.

The correctness probe asks whether the candidate trace asserts the unit claim accurately.
Correctness is therefore stricter than presence: a model must not only mention the relevant
topic, but state the required content correctly. This separation lets us distinguish
coverage failures from content failures. A missing unit indicates that the model did not
attempt a required observation, clinical fact, or inferential bridge; an incorrect unit
indicates that the model engaged with the topic but stated it wrongly.

For each unit $i$, the judge returns a presence label
$p_i \in \{0,1,2\}$ and a correctness label $c_i \in \{-1,0,+1\}$:
\begin{itemize}
    \item \textbf{Presence} $p_i$: $0$ if the unit topic is absent, $1$ if it is mentioned
    only vaguely or implicitly, and $2$ if it is clearly asserted.
    \item \textbf{Correctness} $c_i$: $-1$ if the candidate states the unit incorrectly,
    $0$ if correctness is not applicable because the unit is not meaningfully addressed,
    and $+1$ if the candidate states the unit correctly.
\end{itemize}

Correctness is anchored to the relevant source of truth for each axis. Perception
correctness is judged against the image and its visual evidence. Medical-knowledge
correctness is judged against standard clinical fact. Rationale correctness is judged
against the validity of the inferential move given the premises stated by the candidate
trace. This allows a model to receive credit for a valid reasoning path even when its
surface wording differs from the reference trace.

\paragraph{Axis-level aggregation.}
Let $a \in \{\text{perception}, \text{medical knowledge}, \text{rationale}\}$ denote one
of the three axes, and let $N_a$ be the number of checklist units assigned to axis $a$.
For each axis, we compute a normalized presence mean over all units:
\[
\mathrm{Presence}_a
=
\frac{1}{N_a}
\sum_{i \in a} \frac{p_i}{2}.
\]
This score lies in $[0,1]$ and measures how much of the required reasoning content the
candidate trace attempted to cover.

We compute correctness only over units that the model engaged with (were present):
\[
\mathrm{Correctness}_a
=
\frac{
\left|\{i \in a : p_i \geq 1 \ \wedge \ c_i = +1\}\right|
}{
\left|\{i \in a : p_i \geq 1\}\right|
}.
\]
If no unit in an axis is present, correctness for that axis is treated as not applicable
and the axis receives no correctness credit in the joint score. This conditioning prevents
omitted units from being counted twice: omissions lower the presence score, while
correctness measures the accuracy of the content the model actually attempted.

Each generated trace is therefore summarized by six axis-level scores:
\[
\{
\mathrm{Presence}_a,
\mathrm{Correctness}_a
\}_{a \in
\{\text{perception}, \text{medical knowledge}, \text{rationale}\}}.
\]
These scores diagnose whether a model fails by missing visual evidence, misstating medical
knowledge, or failing to make the required inferential connection to the answer.

\paragraph{Joint trace score.}
When a single headline score is needed, we compute a joint trace score by combining
presence and correctness within each axis:
\[
\mathrm{TraceScore}
=
\frac{1}{2}
\sum_a
\mathrm{Presence}_a
\cdot
\mathrm{Correctness}_a.
\]
The product penalizes both forms of failure: a model receives a low score if it omits
required reasoning content, even when the few units it mentions are correct, and it also
receives a low score if it covers many units but states them incorrectly.

\paragraph{Repeated sampling.}
When a model is sampled multiple times for the same case, we compute the six axis-level
scores and the joint trace score for each generated trace independently. We then report the
mean and sample variance across runs. This lets us measure not only average reasoning
quality, but also the stability of the model's reasoning traces.

\textbf{Held-out split.} \dataset~Bench is constructed from a 1500 held-out question and reasoning partition that does not overlap with \dataset training sources at the source-article, image, or question level.


\begin{promptbox}[title={Prompt for reasoning-unit question generation}, label={prompt:pkr_generation}]
\textbf{Role.}
You are a medical VQA reasoning-evaluation assistant. Your goal is to convert a
reference reasoning trace into a compact checklist of unit questions that can later be
used to evaluate another model's free-form reasoning trace.

\textbf{Task.}
Given a medical visual-question-answering case, the reference answer, and the reference
reasoning trace, generate a non-redundant set of axis-specific unit questions. Each unit
question should capture one atomic component of the reasoning needed to justify the
reference answer.

Generate units along three axes:

\begin{itemize}
    \item \textbf{Perception:} image-grounded observations used by the trace, such as
    modality, anatomy, lesion appearance, spatial pattern, measurement, count, or stated
    absence. Modifiers of the same finding should be merged into one unit.

    \item \textbf{Medical knowledge:} general clinical facts used by the trace. These
    facts should remain true even if the present case is removed.

    \item \textbf{Rationale:} case-specific inferential links that connect the visual
    evidence and medical knowledge to the answer, including diagnosis, differential
    exclusion, mechanism, next step, treatment, prognosis, or severity.
\end{itemize}

For each unit, produce the following fields:

\begin{itemize}
    \item \texttt{topic}: a short neutral noun phrase naming what the unit is about.
    The topic must not contain polarity, values, measurements, direction, or conclusions.

    \item \texttt{claim}: a concise declarative sentence stating the exact proposition
    from the reference trace.

    \item \texttt{presence\_question}: a lenient yes/no question asking whether a
    candidate trace discusses the unit's topic at all.

    \item \texttt{correctness\_question}: a strict yes/no question asking whether the
    candidate trace states the unit's claim correctly.

    \item \texttt{source\_quote}: a verbatim contiguous span from the reference trace that
    supports the unit.

    \item \texttt{importance}: either \texttt{core}, if the unit is necessary to defend the
    final answer, or \texttt{supporting}, otherwise.
\end{itemize}

\textbf{Important distinction.}
The \texttt{topic} and \texttt{claim} must be separated carefully. The topic is a neutral
anchor used to measure coverage. The claim contains the polarity, value, laterality,
measurement, diagnosis, or conclusion. For example, if the reference trace states that
there is no pleural effusion, the topic should be ``pleural effusion,'' while the claim
should be ``There is no pleural effusion.'' This allows a candidate trace that incorrectly
claims an effusion is present to receive presence credit but fail correctness.

\textbf{Rules.}
Do not add information that is not stated in the reference trace. Do not create units that
only restate the question stem or answer options. Do not use option-letter framing; rewrite
option-based explanations as positive medical, visual, or inferential claims. Avoid
duplicates, skip trivial observations, and assign each unit to exactly one axis.

\textbf{Output format.}
Return only a JSON object in the following format:

\begin{verbatim}
{
  "unit_questions": [
    {
      "unit_id": "u1",
      "axis": "perception | medical_knowledge | rationale",
      "topic": "...",
      "claim": "...",
      "presence_question": "...",
      "correctness_question": "...",
      "source_quote": "...",
      "importance": "core | supporting"
    }
  ]
}
\end{verbatim}

\end{promptbox}

\begin{table*}[t]
\centering
\small
\setlength{\tabcolsep}{5pt}
\renewcommand{\arraystretch}{1.15}
\caption{Qualitative comparison of reasoning traces. The image and original question are shown at the top. Perception (P), medical knowledge (K), and rationale (R) units are color-coded; the same colors are used to highlight the corresponding evidence in each model's reasoning. For compactness, we show only the presence questions for each unit.}
\label{tab:reasoning_unit_highlight}

\begin{tabularx}{\textwidth}{@{}X X@{}}
\toprule

\multicolumn{2}{@{}p{\textwidth}@{}}{
\textbf{Example case.}

\vspace{0.5em}

\centering
\includegraphics[width=0.32\textwidth]{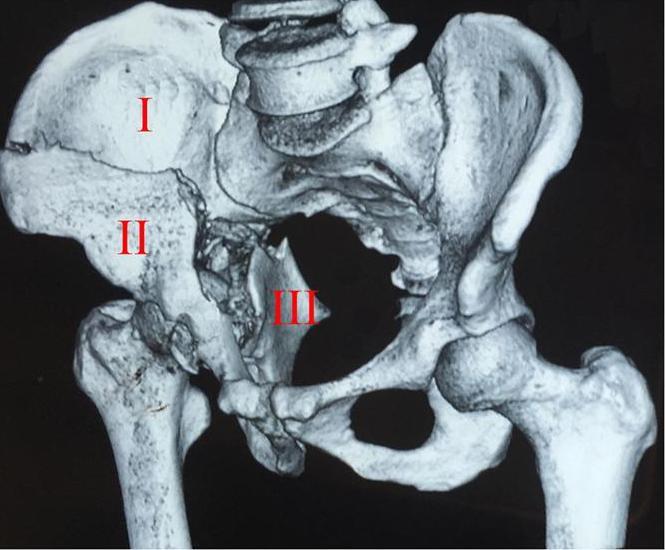}

\vspace{0.5em}

\raggedright
\textbf{Question:} Based on this image, are the findings normal or abnormal?

\vspace{0.3em}

\textbf{Options:} A. Normal \quad B. Abnormal
}
\\

\midrule

\multicolumn{2}{@{}p{\textwidth}@{}}{
\textbf{\dataset-Bench Reasoning Trace Evaluation Checklist.}

\unitone{\textbf{P1:} Does the response discuss bony fragments in the left hemipelvis?}
\quad
\unittwo{\textbf{P2:} Does the response discuss bony discontinuity around the left acetabular/innominate region?}
\quad
\unitthree{\textbf{P3:} Does the response discuss the right hemipelvis appearance?}
\quad
\unitfour{\textbf{K1:} Does the response discuss how normal pelvic 3D reconstruction appears?}
\quad
\unitfive{\textbf{R1:} Does the response link displaced left bony fragments to incompatibility with normal appearance?}
\quad
\unitsix{\textbf{R2:} Does the response conclude that the pelvic findings are abnormal?}
}
\\

\midrule

\textbf{Ours} & \textbf{Qwen2.5-VL-7B} \\
\midrule

\begin{minipage}[t]{0.48\textwidth}
The bones appear \unitone{fragmented} and \unittwo{displaced at these sites}, with irregular bone edges and \unittwo{loss of normal cortical continuity}. The femoral heads and \unittwo{acetabula are involved}, indicating significant trauma or fracture.

\vspace{0.5em}

These features clearly deviate from \unitfour{normal pelvic anatomy, where the bones should be intact, smooth, and aligned}. \unitfive{The presence of such extensive fractures strongly suggests an abnormal finding.}

\vspace{0.5em}

Therefore, the correct answer must indicate \unitsix{abnormality}.
\end{minipage}
&
\begin{minipage}[t]{0.48\textwidth}
The image depicts a pelvic radiograph with labels indicating different regions (I, II, III). These markings are likely used to highlight specific areas for analysis. Without any notable deformities, fractures, or clear signs of pathology noted, the image does not show any obvious abnormalities.

\vspace{0.5em}

In the absence of additional information suggesting abnormality or reference to \unitfour{known normal pelvic anatomy}, the image appears within expected variations that could be considered normal.
\end{minipage}
\\

\bottomrule
\end{tabularx}
\end{table*}

\section{Model Training Details}

\subsection{SFT Training Data}
For supervised fine-tuning, we use a mixed medical vision-language dataset composed of our curated data and several public medical VQA and classification datasets. When a source contains separate open-ended and closed-ended splits, we merge them and report the combined size under the dataset name. Table~\ref{tab:sft_training_data} summarizes the SFT data composition. Our SFT training took 24 hours on 2 H200 GPUs.

\begin{table}[t]
\centering
\small
\setlength{\tabcolsep}{8pt}
\renewcommand{\arraystretch}{1.08}
\caption{Composition of the SFT training data. Open-ended and closed-ended splits are merged when applicable.}
\label{tab:sft_training_data}
\begin{tabular}{l r}
\toprule
\textbf{Data source} & \textbf{\#Examples} \\
\midrule
Ours & 193{,}270 \\
PMC-VQA~\cite{zhang2024development} & 114{,}601 \\
VinDr-CXR~\cite{PhysioNet-vindr-cxr-1.0.0} & 48{,}515 \\
PathVQA~\cite{he-etal-2021-towards} & 33{,}536 \\
HAM10000~\cite{tschandl2018ham10000} & 14{,}334 \\
SLAKE~\cite{liu2021slake} & 10{,}145 \\
VinDr-Mammo~\cite{PhysioNet-vindr-mammo-1.0.0} & 9{,}753 \\
Brain Tumor MRI~\cite{bhuvaji2020brain} & 7{,}378 \\
VQA-RAD~\cite{lau2018dataset} & 3{,}548 \\
Hyper-Kvasir~\cite{borgli2020hyperkvasir} & 2{,}849 \\
EyePACS~\cite{kaggleeyepacs2015diabetic} & 2{,}197 \\
APTOS2019~\cite{aptos2019blindness} & 2{,}080 \\
BUSI~\cite{aldhabyani2020breastultrasound} & 1{,}193 \\
\midrule
\textbf{Total} & \textbf{446{,}399} \\
\bottomrule
\end{tabular}
\end{table}

\subsection{GRPO Training Details}
\label{app:train_detail}

We fine-tune \texttt{Qwen2.5-VL-7} using the GRPO trainer from TRL, initialized from our SFT LoRA adapter (row 3 from Table~\ref{tab:increment}). The model is trained to generate reasoning traces in the structured format
\texttt{<think>...</think><answer>...</answer>}. The reward combines a format component and an answer-correctness component. The format reward gives $+1.0$ for producing exactly one \texttt{<think>} block and $+1.0$ for producing exactly one \texttt{<answer>} block. It applies a $-2.0$ penalty for repeated tags or for placing the answer before the reasoning block, and a $-1.0$ penalty when either tag is opened but not closed, which typically indicates truncation. The correctness reward gives $+2.0$ only when the canonicalized answer exactly matches the ground-truth label, and $0.0$ otherwise. We do not use partial credit or fuzzy matching.

Training is performed on 2 H200 GPUs with \texttt{accelerate} in \texttt{bfloat16} precision. We use a per-device batch size of 1, gradient accumulation of 8, and 8 generations per prompt. The sequence budget is 8192 tokens, with \texttt{max\_prompt\_length=3072} and \texttt{max\_completion\_length=2048}. Optimization uses the \texttt{dr\_grpo} loss with token-level importance sampling, a learning rate of $1\times10^{-6}$, weight decay of 0.1, warmup ratio of 0.1, and gradient clipping at 0.5. Our GRPO training took around 20 hours on 2 H200 GPUs.

To improve training stability, we use several safeguards against the long-completion entropy collapse observed in earlier runs. First, we keep the policy anchored to the SFT prior using a KL coefficient of $\beta=0.04$. Second, we use an asymmetric trust region with \texttt{epsilon=0.20} and \texttt{epsilon\_high=0.28}, limiting the amplification of low-probability tokens from isolated high-reward samples.

\subsubsection{GRPO Training Data Statistics}
\label{app:train_data}

For GRPO, we use a small mixed medical VQA training set sampled from our dataset and four established medical VQA benchmarks. Table~\ref{tab:grpo_training_data} summarizes the number of examples used from each source. All the data used for GRPO is a held out data split from the train split of the datasets that had not been used for SFT.

\begin{table}[t]
\centering
\small
\setlength{\tabcolsep}{8pt}
\renewcommand{\arraystretch}{1.1}
\caption{Composition of the GRPO training data.}
\label{tab:grpo_training_data}
\begin{tabular}{l r}
\toprule
\textbf{Data source} & \textbf{\#Examples} \\
\midrule
\dataset & 3{,}000 \\
PMC-VQA & 2{,}000 \\
SLAKE & 40 \\
RadVQA & 40 \\
PathVQA & 40 \\
\midrule
\textbf{Total} & \textbf{5{,}120} \\
\bottomrule
\end{tabular}
\end{table}

\subsection{Model Evaluation Details}
\label{app:eval_datasets}

To complement the main results, we summarize the evaluation datasets used throughout our experiments in Table~\ref{tab:eval_datasets_appendix}. The benchmark suite covers both medical VQA and image-classification tasks across radiology, pathology, ophthalmology, dermatology, endoscopy, mammography, ultrasound, and general biomedical figures.

\begin{table*}[t]
\centering
\small
\setlength{\tabcolsep}{4pt}
\renewcommand{\arraystretch}{1.12}
\caption{Summary of evaluation datasets used in our experiments. We include both medical VQA benchmarks and image-classification benchmarks, covering radiology, pathology, ophthalmology, dermatology, endoscopy, mammography, ultrasound, and general biomedical figures.}
\label{tab:eval_datasets_appendix}
\begin{tabularx}{\textwidth}{@{} l l X @{}}
\toprule
\textbf{Dataset} & \textbf{Modality / Domain} & \textbf{Description} \\
\midrule
\multicolumn{3}{@{}l}{\textbf{VQA benchmarks}} \\
\midrule
SLAKE & Radiology & Medical VQA benchmark built from radiological images with questions requiring visual recognition and clinical knowledge. \\
VQA-Rad & Radiology & Radiology VQA dataset containing clinician-generated questions over medical images from multiple imaging modalities. \\
PathVQA & Pathology & Pathology VQA benchmark focused on microscopic and histopathological images with open-ended and multiple-choice questions. \\
PMC-VQA & Biomedical figures & Large-scale VQA benchmark constructed from image--caption pairs in PubMed Central articles. \\
MedXpertQA & Multimodal medical QA & Challenging medical VQA benchmark designed to test expert-level medical reasoning over clinical images and questions. \\
JAMA & Clinical medical images & Multimodal clinical QA benchmark based on medical cases and images from JAMA-style clinical content. \\
\midrule
\multicolumn{3}{@{}l}{\textbf{Classification benchmarks}} \\
\midrule
HAM10000 & Dermoscopy & Skin-lesion classification dataset containing dermoscopic images of common pigmented skin lesions. \\
EyePACS & Retinal fundus & Diabetic-retinopathy grading dataset based on color fundus photographs. \\
HyperKvasir & Gastrointestinal endoscopy & Multi-class gastrointestinal endoscopy dataset containing images and videos from the digestive tract. \\
BrainTumorMRI & Brain MRI & MRI classification dataset for distinguishing brain tumor categories from brain scans. \\
VinDr-CXR & Chest X-ray & Chest radiograph dataset for thoracic disease classification and radiological abnormality recognition. \\
VinDr-Mammo & Mammography & Mammography dataset for breast-lesion and cancer-related image classification. \\
BUSI & Breast ultrasound & Breast ultrasound image dataset for classifying normal, benign, and malignant cases. \\
\bottomrule
\end{tabularx}
\end{table*}

\section{Expert review and case analysis}
\label{app:case-review}
\begin{figure}[t]
    \centering

    \begin{subfigure}{0.9\linewidth}
        \centering
        \includegraphics[angle=90,width=\linewidth]{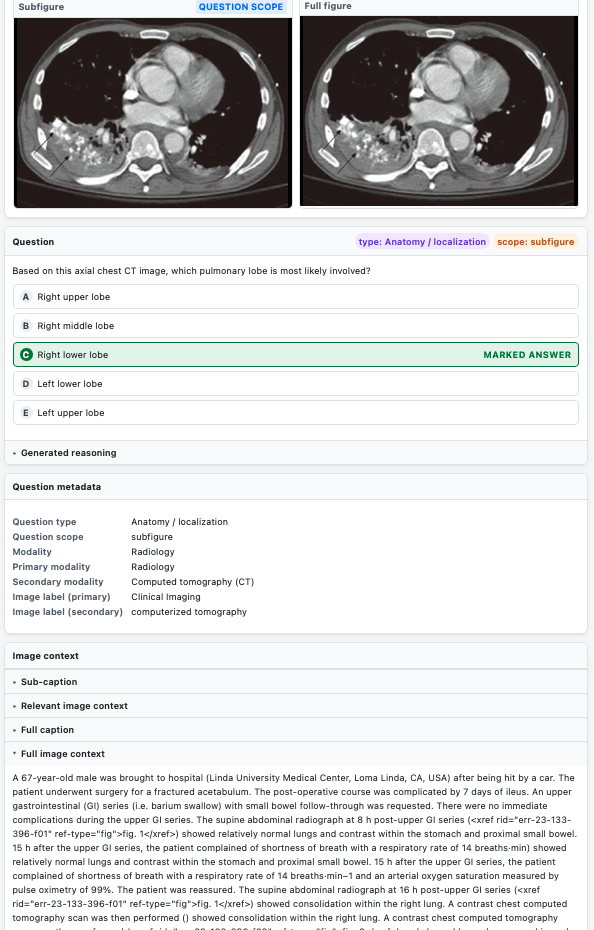}
        \caption{First part.}
        \label{fig:image1}
    \end{subfigure}

    \vspace{0.4em}

    \begin{subfigure}{0.9\linewidth}
        \centering
        \includegraphics[angle=90,width=\linewidth]{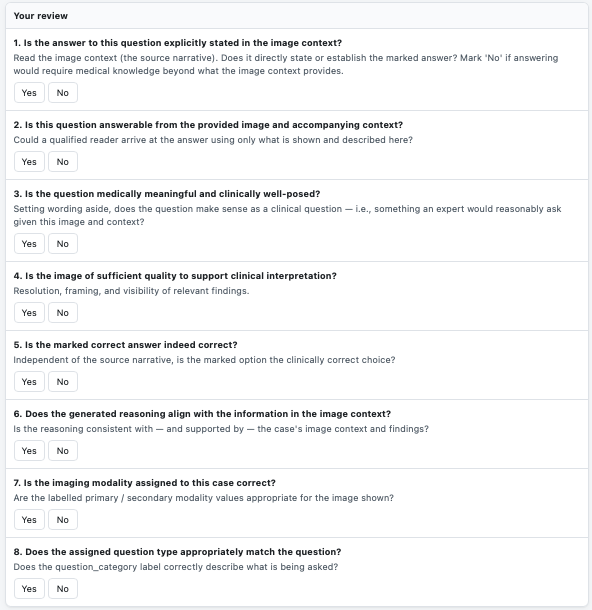}
        \caption{Second part.}
        \label{fig:image2}
    \end{subfigure}

    \caption{A screenshot of the designed framework with 100 examples for the physician expert to review \dataset}
    \label{fig:expert}
\end{figure}

The reviewer was presented with a random sample of $100$ cases drawn from \dataset-Bench using a custom annotation interface (Figure~\ref{fig:expert} that exposes, for each case, the source figure (sub-figure and parent figure), the question stem and option set, the marked correct answer, the case's image context (sub-caption, relevant context, full caption, full image context), the metadata labels (question type, scope, primary and secondary modality), the model-generated reasoning trace, and the
structured \emph{perception}, \emph{medical-knowledge}, and \emph{reasoning} probe sets that drive trace-quality scoring (Section~\ref{sec:reasoning-benchmark}). 
For each case the reviewer answered eleven binary judgements covering
five dimensions of benchmark quality (Appendix~\ref{app:case-review}): (i) \textit{answer correctness}—is
the marked option clinically correct independent of the source narrative?;
(ii) \textit{answerability and grounding}—is the question answerable
from the provided image and context, and is the answer supported by that context?; (iii) \textit{clinical validity}—is the
question something an expert would reasonably ask given the image and
case? and (iv) \textit{label correctness}—are the modality and
question-type labels accurate, and is the image of sufficient quality
to support clinical interpretation?

\clearpage

\section*{Broader Impact and Societal Considerations}
\label{app:impact}
\dataset releases an open dataset of source-grounded medical reasoning
traces and a paired benchmark, with the goal of enabling reproducible
research on medical LVLM reasoning without requiring access to
proprietary models. Two aspects of the design are intentionally
pro-social. First, every reasoning step in
\dataset is anchored to a specific figure and its case-level context,
so downstream errors can be traced back to a concrete piece of evidence
rather than to an opaque model state. Second, because both the dataset and the benchmark are open and run on a
single 7B backbone, academic and clinical research groups can study
what their models attend to, and where they fail, without depending on
frontier closed-source systems.

We are equally explicit about the risks. Models trained on \dataset
are \emph{not} clinical tools and must not be used to inform diagnosis,
treatment, or triage. PubMed Central is a peer-reviewed but biased
corpus: published cases systematically over-represent rare, atypical,
or pedagogically interesting findings relative to their real-world
prevalence, and some modalities are
substantially under-represented relative to radiology and pathology.
Accuracy on \dataset--Bench should therefore not be read as a proxy for
clinical safety or for generalisation across patient populations.
Demographic representation in the underlying figures and captions
inherits whatever disparities exist in the source literature; auditing
derived models for such disparities before any downstream use is
essential. Although all training material is drawn from de-identified,
publicly accessible figures, we recommend that any derived checkpoints
be released with explicit documentation of these limitations, an
intended-use statement that excludes clinical deployment, and a
quantitative summary of modality and demographic coverage so that
downstream users can decide whether the artifact is appropriate for
their setting.


\end{document}